%% file: concrete_distribution_draft.tex
\documentclass{article} % For LaTeX2e
\usepackage{iclr2017_conference, times}
\usepackage{hyperref}
\usepackage{url}

\iclrfinalcopy

\usepackage{commands}
\usepackage{multirow,bigdelim}
\usepackage{booktabs}
\usepackage{textcomp}
\usepackage{adjustbox}
\usepackage{verbatim}
\usepackage{enumitem}
\usepackage{pgfplots}
\usepackage{deepcolor}
\pgfplotsset{compat=newest}
\pgfplotsset{
/pgfplots/colormap={deepblue}{[1cm] rgb255(0cm)=(0,0,0); color(1cm)=(deepblue1); color(2cm)=(deepblue2); color(3cm)=(deepblue3); color(4cm)=(deepblue4); color(5cm)=(deepblue5); color(6cm)=(white)}
}
\usepgfplotslibrary{ternary}
\usepgfplotslibrary{patchplots}
\usepackage{tikz}
\usepgfplotslibrary{external} 
\usepackage{caption}
\usepackage{subcaption}
\usetikzlibrary{calc}
\usetikzlibrary{arrows}
\usepackage{mathtools}

\title{The Concrete Distribution: \\ A Continuous Relaxation of \\ Discrete Random Variables}

\author{Chris J. Maddison\textsuperscript{1,2}, Andriy Mnih\textsuperscript{1}, \& Yee Whye Teh\textsuperscript{1}\\
\textsuperscript{1}DeepMind, London, United Kingdom\\
\textsuperscript{2}University of Oxford, Oxford, United Kingdom\\
\texttt{cmaddis@stats.ox.ac.uk} \\
}

\begin{document}

\maketitle

\input{abstract}

\input{introduction}

\input{background}

\input{concrete}

\input{related}

\input{experiments}

\input{conclusion}

\subsubsection*{Acknowledgments}
We thank Jimmy Ba for the excitement and ideas in the early days, Stefano Favarro for some analysis of the distribution. We also thank Gabriel Barth-Maron and Roger Grosse.

\vspace{-\baselineskip}

{
\setlength{\bibsep}{2.5pt}
\bibliography{concrete_distribution}
\bibliographystyle{iclr2017_conference}
}

\appendix
\input{appendix}

\end{document}

%% file: abstract.tex
\begin{abstract}
The reparameterization trick enables optimizing large scale stochastic computation graphs via gradient descent. The essence of the trick is to refactor each stochastic node into a differentiable function of its parameters and a random variable with fixed distribution. After refactoring, the gradients of the loss propagated by the chain rule through the graph are low variance unbiased estimators of the gradients of the expected loss. While many continuous random variables have such reparameterizations, discrete random variables lack useful reparameterizations due to the discontinuous nature of discrete states. In this work we introduce {\sc Concrete} random variables---{\sc con}tinuous relaxations of dis{\sc crete} random variables. The Concrete distribution is a new family of distributions with closed form densities and a simple reparameterization. Whenever a discrete stochastic node of a computation graph can be refactored into a one-hot bit representation that is treated continuously, Concrete stochastic nodes can be used with automatic differentiation to produce low-variance biased gradients of objectives (including objectives that depend on the log-probability of latent stochastic nodes) on the corresponding discrete graph. We demonstrate the effectiveness of Concrete relaxations on density estimation and structured prediction tasks using neural networks.
\end{abstract}

%% file: introduction.tex
\section{Introduction}
\label{sec:introduction}

Software libraries for automatic differentiation (AD) \citep{tensorflow2015-whitepaper, theano} are enjoying broad use, spurred on by the success of neural networks on some of the most challenging problems of machine learning. The dominant mode of development in these libraries is to define a forward parametric computation, in the form of a directed acyclic graph, that computes the desired objective. If the components of the graph are differentiable, then a backwards computation for the gradient of the objective can be derived automatically with the chain rule. The ease of use and unreasonable effectiveness of gradient descent has led to an explosion in the diversity of architectures and objective functions. Thus, expanding the range of useful continuous operations can have an outsized impact on the development of new models. For example, a topic of recent attention has been the optimization of stochastic computation graphs from samples of their states. Here, the observation that AD ``just works'' when stochastic nodes\footnote{For our purposes a stochastic node of a computation graph is just a random variable whose distribution depends in some deterministic way on the values of the parent nodes.} can be reparameterized into deterministic functions of their parameters and a fixed noise distribution \citep{kingma2013auto, rezende2014stochastic}, has liberated researchers in the development of large complex stochastic architectures \citep[\eg][]{gregor2015draw}. 

Computing with discrete stochastic nodes still poses a significant challenge for AD libraries. Deterministic discreteness can be relaxed and approximated reasonably well with sigmoidal functions or the softmax \citep[see e.g.,][]{grefenstette2015learning, graves2016hybrid}, but, if a distribution over discrete states is needed, there is no clear solution. There are well known unbiased estimators for the gradients of the parameters of a discrete stochastic node from samples. While these can be made to work with AD, they involve special casing and defining surrogate objectives \citep{schulman2015gradient}, and even then they can have high variance. Still, reasoning about discrete computation comes naturally to humans, and so, despite the difficulty associated, many modern architectures incorporate discrete stochasticity \citep{mnih2014attention, xu2015show, kocisky2016parsing}. 

This work is inspired by the observation that many architectures treat discrete nodes continuously, and gradients rich with counterfactual information are available for each of their possible states. We introduce a {\sc con}tinuous relaxation of dis{\sc crete} random variables, {\sc Concrete} for short, which allow gradients to flow through their states. The {\it Concrete distribution} is a new parametric family of continuous distributions on the simplex with closed form densities. Sampling from the Concrete distribution is as simple as taking the softmax of logits perturbed by fixed additive noise. This reparameterization means that Concrete stochastic nodes are quick to implement in a way that ``just works'' with AD. Crucially, every discrete random variable corresponds to the zero temperature limit of a Concrete one. In this view optimizing an objective over an architecture with discrete stochastic nodes can be accomplished by gradient descent on the samples of the corresponding Concrete relaxation. When the objective depends, as in variational inference, on the log-probability of discrete nodes, the Concrete density is used during training in place of the discrete mass. At test time, the graph with discrete nodes is evaluated.

The paper is organized as follows. We provide a background on stochastic computation graphs and their optimization in Section \ref{sec:background}. Section \ref{sec:Concrete} reviews a reparameterization for discrete random variables, introduces the Concrete distribution, and discusses its application as a relaxation. Section \ref{sec:related} reviews related work. In Section \ref{sec:experiments} we present results on a density estimation task and a structured prediction task on the MNIST and Omniglot datasets. In Appendices \ref{appendix:Concreterelaxations} and \ref{appendix:cheatsheet} we provide details on the practical implementation and use of Concrete random variables. When comparing the effectiveness of gradients obtained via Concrete relaxations to a state-of-the-art-method \citep[VIMCO,][]{MnihRezende2016}, we find that they are competitive---occasionally outperforming and occasionally underperforming---all the while being implemented in an AD library without special casing.

%% file: background.tex
\section{Background}
\label{sec:background}

\subsection{Optimizing Stochastic Computation Graphs}
\label{sec:scgs}

Stochastic computation graphs (SCGs) provide a formalism for
specifying input-output mappings, potentially stochastic, with learnable
parameters using directed acyclic graphs (see \citet{schulman2015gradient} for a review).  The state of each non-input node in such a graph is obtained from the states of its parent nodes by either
evaluating a deterministic function or sampling from a conditional
distribution.  Many training objectives in supervised, unsupervised, and
reinforcement learning can be expressed in terms of SCGs.

To optimize an objective represented as a SCG, we need estimates of its parameter gradients. We will
concentrate on graphs with some stochastic nodes (backpropagation covers the rest). For simplicity, we restrict our attention to graphs with a single stochastic node $X$.
We can interpret the forward pass in the graph as first sampling
$X$ from the conditional distribution $p_{\phi}(x)$ of the stochastic node given
its parents, then evaluating a deterministic function $f_{\theta}(x)$ at $X$. We can think of $f_{\theta}(X)$ as a noisy objective, and we are interested in optimizing its expected value $L(\theta, \phi)=\mathbb{E}_{X \sim p_{\phi}(x)}[f_{\theta}(X)]$ \wrt parameters $\theta, \phi$.

In general, both the objective and its gradients are intractable. We will side-step this issue by estimating them with samples from $p_{\phi}(x)$. The gradient \wrt to the parameters $\theta$ 
has the form
\begin{align}
\label{eqn:grad_f}
    \ddTheta L(\theta, \phi) = \ddTheta \mathbb{E}_{X \sim p_{\phi}(x)}[f_{\theta}(X)] = \mathbb{E}_{X \sim p_{\phi}(x)}[\ddTheta f_{\theta}(X)]
\end{align}
and can be easily estimated using Monte Carlo sampling:
\begin{align}
    \ddTheta L(\theta, \phi) \simeq \frac{1}{S} \sum\nolimits_{s=1}^{S} \ddTheta f_{\theta}(X^s),
\end{align}
where $X^s \sim p_{\phi}(x)$ i.i.d.
The more challenging task is to compute the gradient \wrt
the parameters $\phi$ of $p_{\phi}(x)$. The expression obtained by differentiating
the expected objective,
\begin{align}
\label{eqn:grad_p_naive}
    \ddPhi L(\theta, \phi) =  \ddPhi \int \! p_{\phi}(x) f_{\theta}(x) \, \mathrm{d}x = \int \! f_{\theta}(x) \ddPhi p_{\phi}(x) \, \mathrm{d}x,
\end{align}
does not have the form of an expectation \wrt $x$ and thus does not directly lead
to a Monte Carlo gradient estimator. However, there are two ways of getting around
this difficulty which lead to the two classes of estimators we will now discuss.

\subsection{Score Function Estimators}

The \emph{score function estimator} \citep[SFE,][]{fu2006gradient}, also known as the REINFORCE \citep{williams1992simple} or likelihood-ratio estimator \citep{glynn1990likelihood}, is based on the identity
$\ddPhi p_{\phi}(x) = p_{\phi}(x) \ddPhi \log p_{\phi}(x)$, which allows the gradient in \eqn{eqn:grad_p_naive} to be written as
an expectation:
\begin{align}
    \ddPhi L(\theta, \phi) = \mathbb{E}_{X \sim p_\phi(x)}\left[f_{\theta}(X) \ddPhi \log p_\phi(X)\right].
\end{align}
Estimating this expectation using naive Monte Carlo gives the 
estimator
\begin{align}
    \ddPhi L(\theta, \phi) \simeq \frac{1}{S} \sum\nolimits_{s=1}^{S} f_{\theta}(X^s) \ddPhi \log p_\phi(X^s),
\end{align}
where $X^s \sim p_\phi(x)$ i.i.d.
This is a very general estimator that is applicable whenever $\log p_\phi(x)$
is differentiable \wrt $\phi$. As it does not require $f_{\theta}(x)$ to be
differentiable or even continuous as a function of $x$, the SFE can be used with
both discrete and continuous random variables.

Though the basic version of the estimator can suffer from high variance,
various variance reduction techniques can be used to make the estimator much
more effective \citep{GreensmithBB04}. Baselines are the most important and
widely used of these techniques \citep{williams1992simple}. A number of score function
estimators have been developed in machine learning \citep{PaisleyBJ12, gregor2013deep, ranganath2014black,
MnihGregor2014, titsias2015local, gu2016muprop}, which differ primarily in the
variance reduction techniques used. 

\subsection{Reparameterization Trick}
\label{subsec:reparam}
In many cases we can sample from $p_{\phi}(x)$ by first sampling $Z$ from some fixed
distribution $q(z)$ and then transforming the sample using some function $g_{\phi}(z)$.
For example, a sample from $\Normal(\mu,\sigma^2)$ 
can be obtained by sampling $Z$ from the standard form of the distribution $\Normal(0,1)$ 
and then transforming it using $g_{\mu, \sigma}(Z) = \mu + \sigma Z$.
This two-stage reformulation of the sampling process, called the \textit{reparameterization
trick}, allows us to transfer the dependence
on $\phi$ from $p$ into $f$ by writing $f_{\theta}(x) = f_{\theta}(g_{\phi}(z))$ for $x = g_{\phi}(z)$,
making it possible to reduce the problem of estimating the gradient \wrt parameters of
a distribution to the simpler problem of estimating the gradient \wrt parameters
of a deterministic function.

Having reparameterized $p_{\phi}(x)$, we can now
express the objective as an expectation \wrt $q(z)$:
\begin{align}
    L(\theta, \phi) = \mathbb{E}_{X \sim p_{\phi}(x)}[f_{\theta}(X)] = \mathbb{E}_{Z \sim q(z)}[f_{\theta}(g_{\phi}(Z))].
\end{align}
As $q(z)$ does not depend on $\phi$, we can estimate the gradient \wrt $\phi$
in exactly the same way we estimated the gradient \wrt $\theta$ in \eqn{eqn:grad_f}.
Assuming differentiability of $f_{\theta}(x)$ \wrt $x$ and of $g_{\phi}(z)$ 
\wrt $\phi$ and using the chain rule gives
\begin{align}
    \ddPhi L(\theta, \phi) = \mathbb{E}_{Z \sim q(z)}[\ddPhi f_{\theta}(g_{\phi}(Z))] = \mathbb{E}_{Z \sim q(z)}\left [f_{\theta}'(g_{\phi}(Z)) \ddPhi g_{\phi}(Z)\right].
\end{align}

The reparameterization trick, introduced in the context of variational
inference independently by \citet{kingma2014auto},
\citet{rezende2014stochastic}, and \citet{titsias14}, is usually the estimator of choice when it is applicable. For continuous latent variables which are not directly reparameterizable, new hybrid estimators have also been developed, by combining partial reparameterizations with score function estimators \citep{ruiz2016generalized,
naesseth2016rejection}.

\subsection{Application: Variational Training of Latent Variable Models}
\label{sec:lvm_training}

We will now see how the task of training latent variable models can be
formulated in the SCG framework. Such models assume that each observation $x$
is obtained by first sampling a vector of latent variables $Z$ from the prior
$p_{\theta}(z)$ before sampling the observation itself from $p_{\theta}(x
\given z)$. Thus the probability of observation $x$ is $p_{\theta}(x) =  \sum_{z}
p_{\theta}(z)p_{\theta}(x \given z)$. Maximum likelihood training of such models
is infeasible, because the log-likelihood (LL) objective
$L(\theta) = \log p_{\theta}(x) = \log \mathbb{E}_{Z \sim p_{\theta}(z)}[p_{\theta}(x \given Z)]$ is
typically intractable and does not fit into the above framework
due to the expectation being inside the $\log$. The multi-sample variational
objective \citep{burda2015importance},
\begin{align}
\label{eqn:iw_objective}
\mathcal{L}_m(\theta, \phi) = \expect_{Z^i \sim q_{\phi}(z | x) }\left[ \log\left( \frac{1}{m} \sum_{i=1}^m \frac{p_{\theta}(Z^i, \  x)}{q_{\phi}(Z^i \given x)} \right)\right].
\end{align}
provides a convenient alternative which has precisely the form we considered in
\sect{sec:scgs}. This approach relies on introducing an auxiliary distribution
$q_{\phi}(z \given x)$ with its own parameters, which serves as approximation
to the intractable posterior $p_{\theta}(z \given x)$. The model is trained by
jointly maximizing the objective \wrt to the parameters of $p$ and $q$.  The
number of samples used inside the objective $m$ allows trading off the
computational cost against the tightness of the bound.  For $m=1$,
$\mathcal{L}_m(\theta, \phi)$ becomes is the widely used evidence lower bound
\citep[ELBO,][]{hoffman2013stochastic} on $\log p_{\theta}(x)$, while for $m > 1$, it is known as the
importance weighted bound \citep{burda2015importance}.

%% file: concrete.tex
\section{The Concrete Distribution}
\label{sec:Concrete}

\subsection{Discrete Random Variables and the Gumbel-Max Trick}

To motivate the construction of Concrete random variables, we review a method for sampling from discrete distributions called the Gumbel-Max trick \citep{luce1959individual, yellott1977relationship, papandreou2011perturb, hazan2012partition, maddison2014astarsamp}. We restrict ourselves to a representation of discrete states as vectors $d \in \{0,1\}^n$ of bits that are one-hot, or $\sum_{k=1}^n d_k = 1$. This is a flexible representation in a computation graph; to achieve an integral representation take the inner product of $d$ with $(1, \ldots, n)$, and to achieve a point mass representation in $\R^m$ take $Wd$ where $W \in \R^{m \times n}$. 

\begin{figure}[t]
\vspace{-\baselineskip}
\begin{center}
     \begin{subfigure}{0.48\textwidth}
          \centering
          \resizebox{0.75\linewidth}{!}{\includegraphics{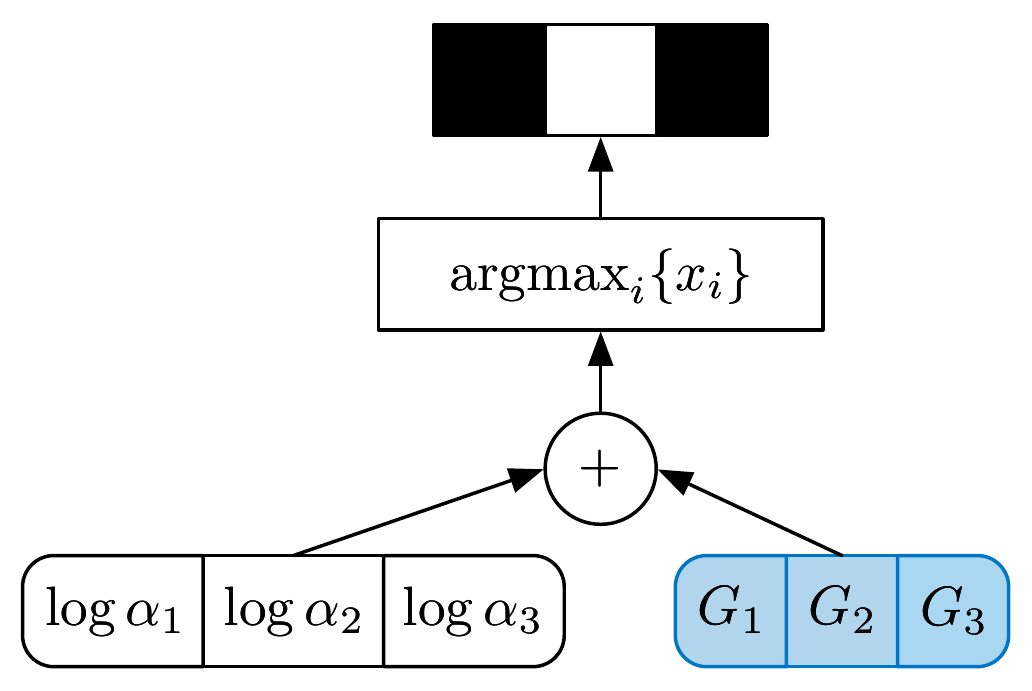}}
          \caption{$\Discrete(\alpha)$}
	\label{subfig:discrete}
     \end{subfigure}
     \begin{subfigure}{0.48\textwidth}
          \centering
          \resizebox{0.75\linewidth}{!}{\includegraphics{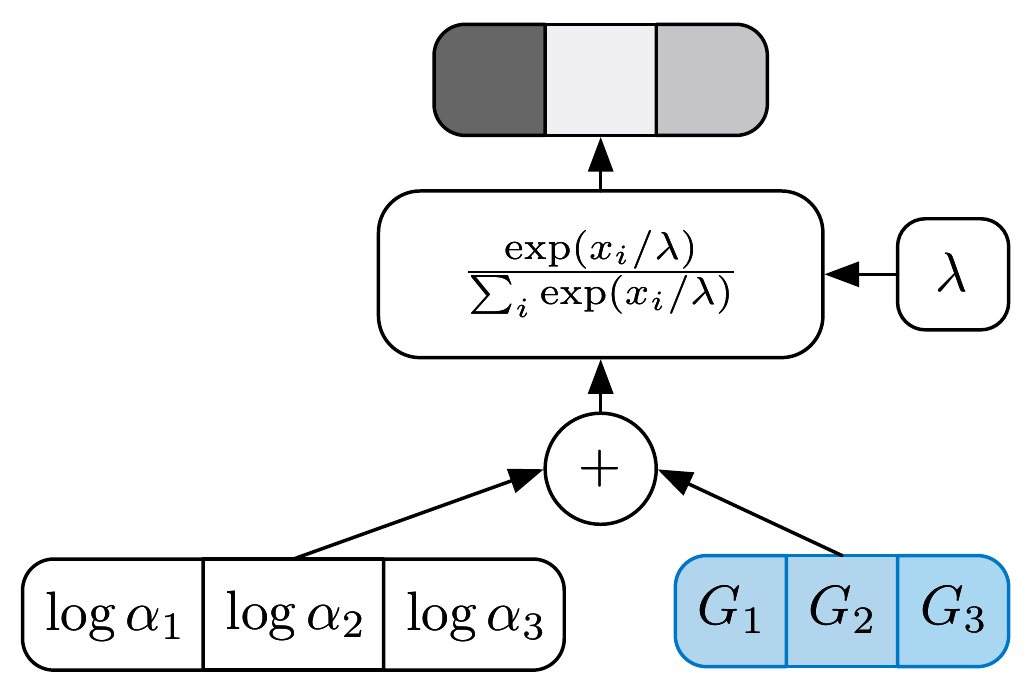}}
          \caption{$\Concrete(\alpha, \lambda)$}
	\label{subfig:Concrete}
     \end{subfigure}
\end{center}
\caption{Visualization of sampling graphs for 3-ary discrete $D \sim \Discrete(\alpha)$ and 3-ary Concrete $X\sim \Concrete(\alpha, \lambda)$. White operations are deterministic, blue are stochastic, rounded are continuous, square discrete. The top node is an example state; brightness indicates a value in [0,1].}
\label{fig:gmt}
\vspace{-\baselineskip}
\end{figure}

Consider an unnormalized parameterization $(\alpha_1, \ldots, \alpha_n)$ where $\alpha_k \in (0, \infty)$ of a discrete distribution $D \sim \Discrete(\alpha)$---we can assume that states with 0 probability are excluded. The Gumbel-Max trick proceeds as follows: sample $U_k \sim \Uniform(0,1)$ i.i.d. for each $k$, find $k$ that maximizes $\{\log \alpha_k  - \log(-\log U_k)\}$, set $D_k = 1$ and the remaining $D_i = 0$ for $i \neq k$. Then
\begin{align}
\label{eq:gmt}
\proba(D_k = 1) = \frac{\alpha_k}{\sum_{i=1}^n \alpha_i}.
\end{align}
In other words, the sampling of a discrete random variable can be refactored into a deterministic function---componentwise addition followed by $\argmax$---of the parameters $\log \alpha_k$ and fixed distribution $-\log(-\log U_k)$. See Figure \ref{subfig:discrete} for a visualization.

The apparently arbitrary choice of noise gives the trick its name, as $-\log(-\log U)$ has a Gumbel distribution. This distribution features in extreme value theory \citep{gumbel1954statistical} where it plays a central role similar to the Normal distribution: the Gumbel distribution is stable under $\max$ operations, and for some distributions, the order statistics (suitably normalized) of i.i.d. draws approach the Gumbel in distribution. The Gumbel can also be recognized as a $-\log$-transformed exponential random variable. So, the correctness of (\ref{eq:gmt}) also reduces to a well known result regarding the $\argmin$ of exponential random variables. See \citep{hazan2016perturbation} for a collection of related work, and particularly the chapter \citep{maddison2016poisson} for a proof and generalization of this trick.

\subsection{Concrete Random Variables}
The derivative of the $\argmax$ is 0 everywhere except at the boundary of state changes, where it is undefined. For this reason the Gumbel-Max trick is not a suitable reparameterization for use in SCGs with AD. Here we introduce the Concrete distribution motivated by considering a graph, which is the same as Figure \ref{subfig:discrete} up to a continuous relaxation of the $\argmax$ computation, see Figure \ref{subfig:Concrete}. This will ultimately allow the optimization of parameters $\alpha_k$ via gradients. 

\begin{figure}[t]
\vspace{-\baselineskip}
\begin{center}
     \begin{subfigure}{0.24\textwidth}
          \centering
          \resizebox{\linewidth}{!}{\includegraphics{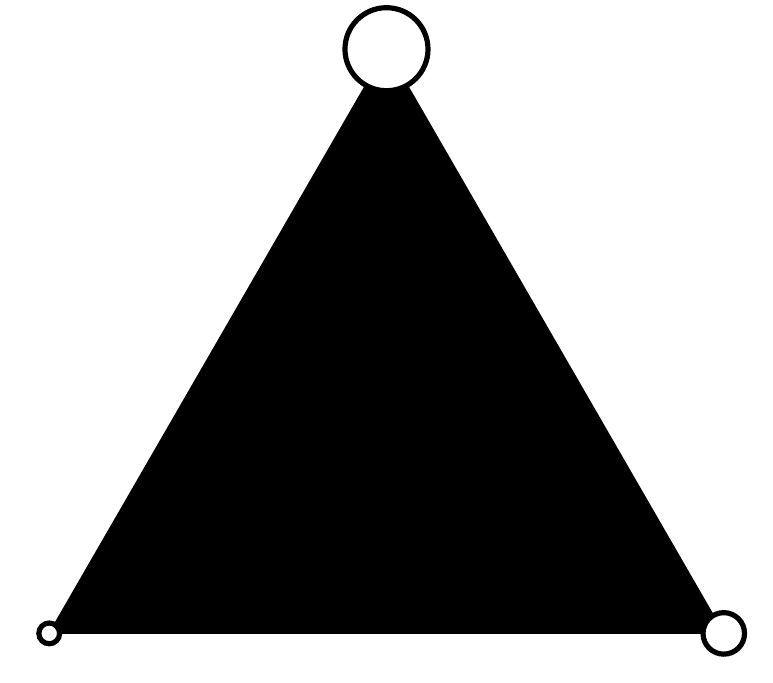}}
          \caption{$\lambda=0$}
     \end{subfigure}
     \begin{subfigure}{0.24\textwidth}
          \centering
          \resizebox{\linewidth}{!}{\includegraphics{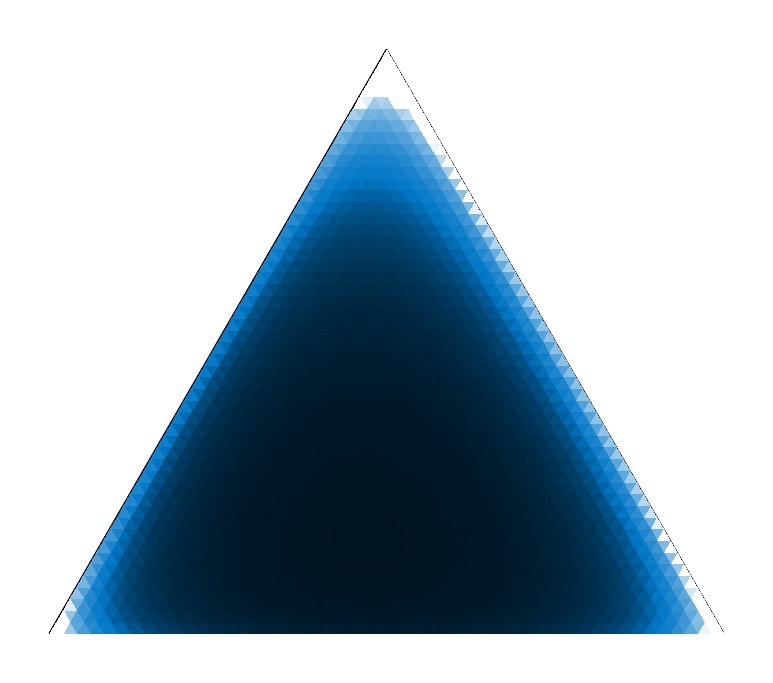}}
          \caption{$\lambda=1/2$}
     \end{subfigure}
     \begin{subfigure}{0.24\textwidth}
          \centering
          \resizebox{\linewidth}{!}{\includegraphics{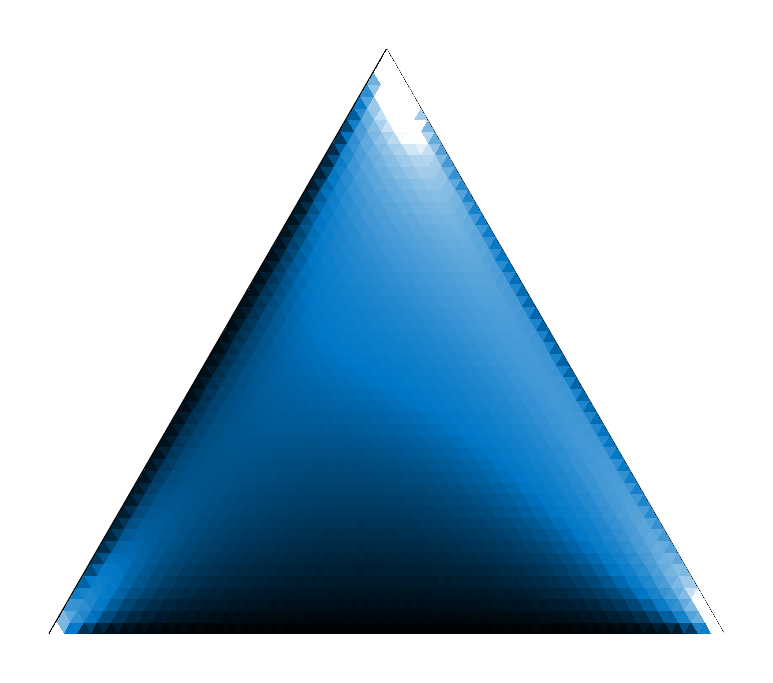}}
          \caption{$\lambda=1$}
     \end{subfigure}
     \begin{subfigure}{0.24\textwidth}
          \centering
          \resizebox{\linewidth}{!}{\includegraphics{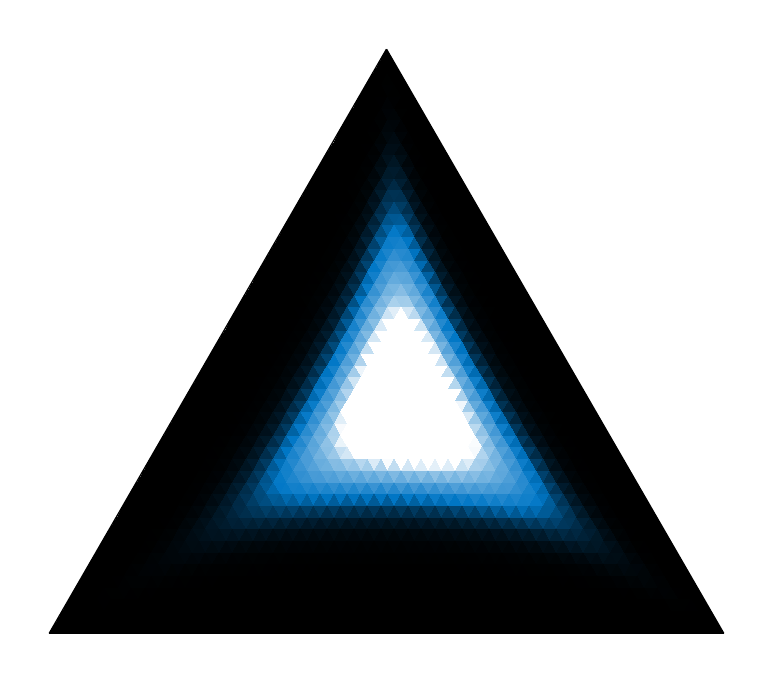}}
          \caption{$\lambda=2$}
          \label{subfig:hightemp}
     \end{subfigure}
\end{center}
\caption{A discrete distribution with unnormalized probabilities $(\alpha_1, \alpha_2, \alpha_3) = (2, 0.5, 1)$ and three corresponding Concrete densities at increasing temperatures $\lambda$. Each triangle represents the set of points $(y_1, y_2, y_3)$ in the simplex $\Delta^2 = \{(y_1, y_2, y_3) \given y_k \in (0,1), y_1 + y_2 + y_3 = 1\}$. For $\lambda = 0$ the size of white circles represents the mass assigned to each vertex of the simplex under the discrete distribution. For $\lambda \in \{2, 1, 0.5\}$ the intensity of the shading represents the value of $p_{\alpha, \lambda}(y)$.}
          \label{fig:simplex}
          \vspace{-\baselineskip}
\end{figure}

The $\argmax$ computation returns states on the vertices of the simplex $\Delta^{n-1} = \{x \in \R^n \given x_k \in [0,1], \sum_{k=1}^n x_k = 1 \}$. The idea behind Concrete random variables is to relax the state of a discrete variable from the vertices into the interior where it is a random probability vector---a vector of numbers between 0 and 1 that sum to 1. To sample a Concrete random variable $X \in \Delta^{n-1}$ at temperature $\lambda \in (0, \infty)$ with parameters $\alpha_k \in (0, \infty)$, sample $G_k \sim \Gumbel$ i.i.d. and set
\begin{align}
\label{eq:sample_Concrete}
X_k = \frac{\exp((\log \alpha_k + G_k)/\lambda)}{\sum_{i=1}^n \exp((\log \alpha_i + G_i)/\lambda)}.
\end{align}
The softmax computation of (\ref{eq:sample_Concrete}) smoothly approaches the discrete $\argmax$ computation as $\lambda \to 0$ while preserving the relative order of the Gumbels $\log \alpha_k + G_k$. So, imagine making a series of forward passes on the graphs of Figure \ref{fig:gmt}. Both graphs return a stochastic value for each forward pass, but for smaller temperatures the outputs of Figure \ref{subfig:Concrete} become more discrete and eventually indistinguishable from a typical forward pass of Figure \ref{subfig:discrete}.

The distribution of $X$ sampled via (\ref{eq:sample_Concrete}) has a closed form density on the simplex. Because there may be other ways to sample a Concrete random variable, we take the density to be its definition.
\begin{defn}[Concrete Random Variables]
\label{def:Concrete}
Let $\alpha \in (0, \infty)^n$ and $\lambda \in (0, \infty)$. $X \in \Delta^{n-1}$ has a Concrete distribution $X \sim \Concrete(\alpha, \lambda)$ with location $\alpha$ and temperature $\lambda$, if its density is:
\begin{align}
\label{eq:condense}
p_{\alpha, \lambda}(x) =(n-1)!  \  \lambda^{n-1} \prod_{k=1}^n \left(\frac{\alpha_k x_k^{-\lambda - 1}}{\sum_{i=1}^n \alpha_i x_i^{-\lambda }}\right).
\end{align}
\end{defn}

Proposition \ref{prop:conc} lists a few properties of the Concrete distribution. \ref{itm:reparam} is confirmation that our definition corresponds to the sampling routine (\ref{eq:sample_Concrete}). \ref{itm:rounding} confirms that rounding a Concrete random variable results in the discrete random variable whose distribution is described by the logits $\log \alpha_k$, \ref{itm:zerotemp} confirms that taking the zero temperature limit of a Concrete random variable is the same as rounding. Finally, \ref{itm:convex} is a convexity result on the density. We prove these results in Appendix \ref{appendix:prop1}. 

\begin{prop}[Some Properties of Concrete Random Variables]
\label{prop:conc}
Let $X  \sim \Concrete(\alpha, \lambda)$ with location parameters $\alpha \in (0, \infty)^n$ and temperature $\lambda \in (0, \infty)$, then
\begin{enumerate}[label=(\alph*)]
\item\label{itm:reparam}  (Reparameterization) If $G_k \sim \Gumbel$ i.i.d., then $X_k \overset{d}{=} \frac{\exp((\log \alpha_k + G_k)/\lambda)}{\sum_{i=1}^n \exp((\log \alpha_i + G_i)/\lambda)}$,
\item\label{itm:rounding}  (Rounding) $\proba\left(X_k > X_i\  \text{ for } i \neq k\right) = \alpha_k / (\sum_{i=1}^n \alpha_i)$,
\item\label{itm:zerotemp}  (Zero temperature) $\proba\left(\lim_{\lambda \to 0} X_k = 1\right) = \alpha_k / (\sum_{i=1}^n \alpha_i)$,
\item\label{itm:convex} (Convex eventually) If $\lambda \leq (n-1)^{-1}$, then $p_{\alpha, \lambda}(x)$ is log-convex in $x$.
\end{enumerate}
\end{prop}
The binary case of the Gumbel-Max trick simplifies to passing additive noise through a step function. The corresponding Concrete relaxation is implemented by passing additive noise through a sigmoid---see Figure \ref{fig:binConcrete}. We cover this more thoroughly in Appendix \ref{appendix:bingumbelmax}, along with a cheat sheet (Appendix \ref{appendix:cheatsheet}) on the density and implementation of all the random variables discussed in this work.

\begin{figure}[t]
\vspace{-\baselineskip}
\begin{center}
\begin{subfigure}{.24\textwidth}
          \resizebox{\linewidth}{!}{\includegraphics{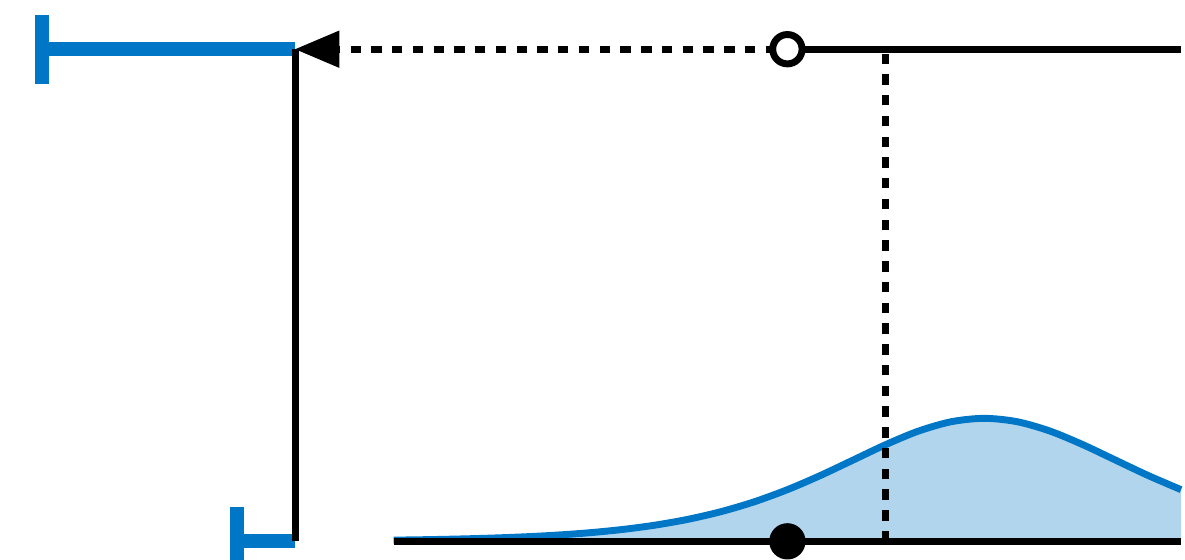}}
\caption{$\lambda = 0$}
\end{subfigure}\hfill
\begin{subfigure}{.24\textwidth}
          \resizebox{\linewidth}{!}{\includegraphics{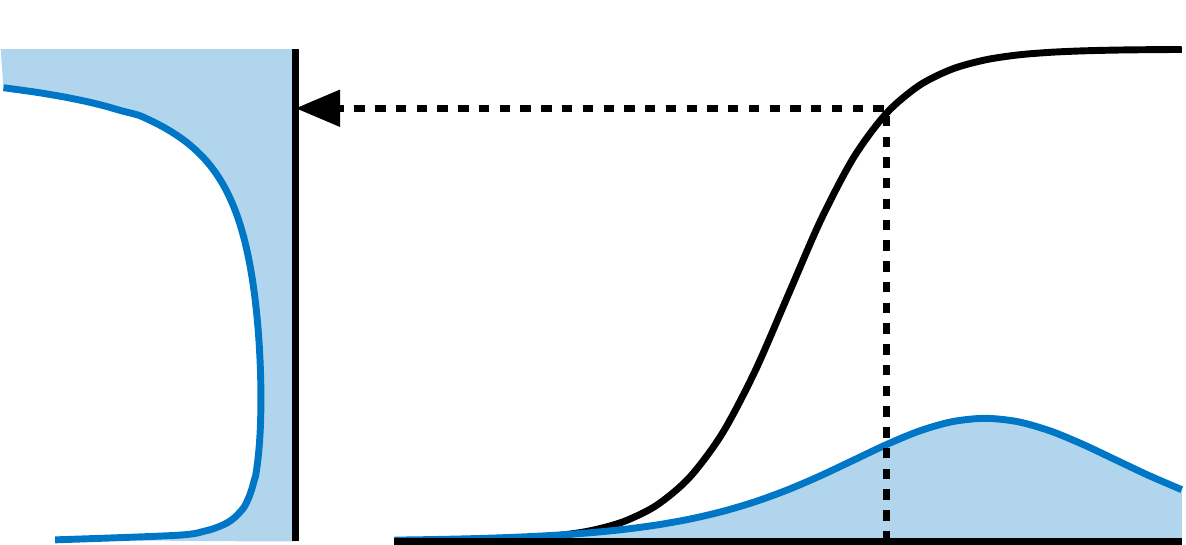}}
\caption{$\lambda = 1/2$}
\label{subfig:binconvex}
\end{subfigure}
\begin{subfigure}{.24\textwidth}
          \resizebox{\linewidth}{!}{\includegraphics{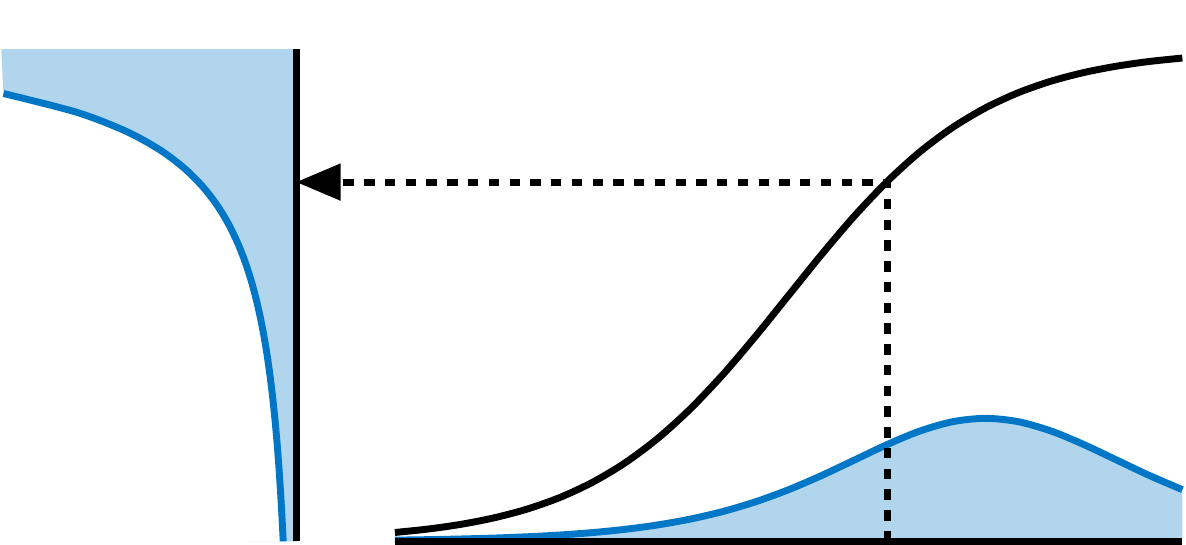}}
\caption{$\lambda = 1$}
\end{subfigure}\hfill
\begin{subfigure}{.24\textwidth}
          \resizebox{\linewidth}{!}{\includegraphics{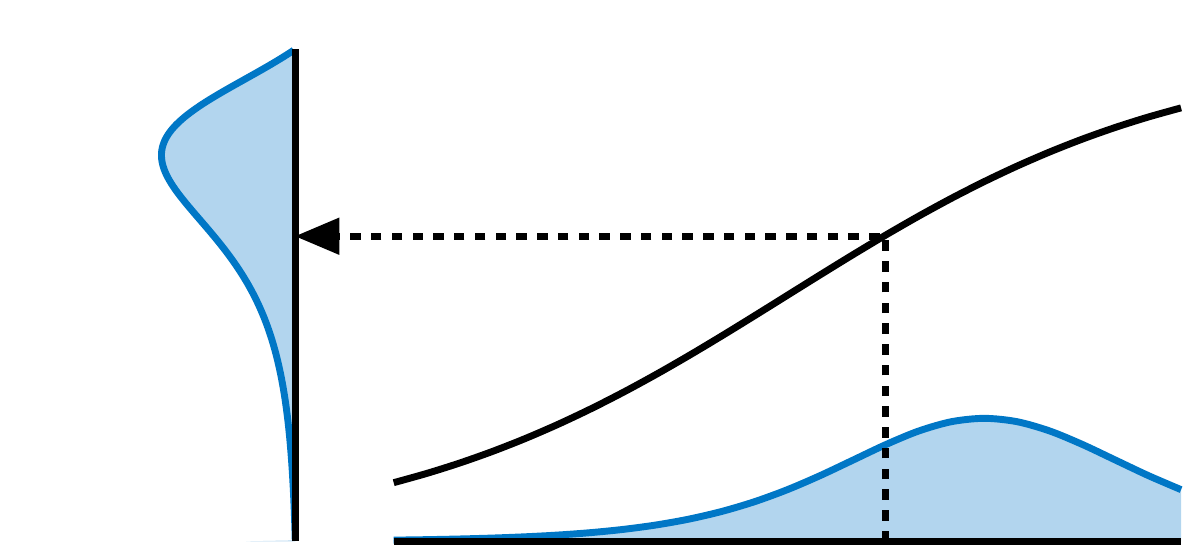}}
\caption{$\lambda = 2$}
\label{subfig:mode}
\end{subfigure}
\end{center}
\caption{A visualization of the binary special case. (a) shows the discrete trick, which works by passing a noisy logit through the unit step function. (b), (c), (d) show Concrete relaxations; the horizontal blue densities show the density of the input distribution and the vertical densities show the corresponding Binary Concrete density on $(0,1)$ for varying $\lambda$.}
\label{fig:binConcrete}
\vspace{-\baselineskip}
\end{figure}

\subsection{Concrete Relaxations}
Concrete random variables may have some intrinsic value, but we investigate them simply as surrogates for optimizing a SCG with discrete nodes. When it is computationally feasible to integrate over the discreteness, that will always be a better choice. Thus, we consider the use case of optimizing a large graph with discrete stochastic nodes from samples. 

First, we outline our proposal for how to use Concrete relaxations by considering a variational autoencoder with a single discrete latent variable. Let $P_a(d)$ be the mass function of some $n$-dimensional one-hot discrete random variable with unnormalized probabilities $a \in (0, \infty)^n$ and $p_{\theta}(x|d)$ some distribution over a data point $x$ given $d \in (0,1)^n$ one-hot. The generative model is then $p_{\theta, a}(x, d) = p_{\theta}(x | d)P_a(d)$. Let $Q_{\alpha}(d | x)$ be an approximating posterior over $d \in (0,1)^n$ one-hot whose unnormalized probabilities $\alpha(x) \in (0,\infty)^n$ depend on $x$. All together the variational lowerbound  we care about stochastically optimizing is
\begin{align}
\label{eq:concrelaxexample}
\mathcal{L}_1(\theta, a, \alpha) = \expect_{D \sim Q_{\alpha}(d | x)}\left[\log \frac{p_{\theta}(x | D) P_{a}(D)}{Q_{\alpha}(D | x)}\right],
\end{align}
with respect to $\theta$, $a$, and any parameters of $\alpha$. First, we relax the stochastic computation $D \sim \Discrete(\alpha(x))$ by replacing $D$ with a Concrete random variable $Z \sim \Concrete(\alpha(x), \lambda_1)$ with density $q_{\alpha, \lambda_1}(z|x)$. Simply replacing every instance of $D$ with $Z$ in Eq. \ref{eq:concrelaxexample} will result in a non-interpretable objective, which does not necessarily lowerbound $\log p(x)$, because $\expect_{Z \sim q_{\alpha, \lambda_1}(a | x)}[-\log Q_{\alpha}(Z | x) / P_a(Z)]$ is not a KL divergence. Thus we propose ``relaxing'' the terms $P_a(d)$ and $Q_{\alpha}(d | x)$ to reflect the true sampling distribution. Thus, the relaxed objective is:
\begin{align}
\label{eq:concrelaxexample2}
\mathcal{L}_1(\theta, a, \alpha) \overset{\text{relax}}{\leadsto} \expect_{Z \sim q_{\alpha, \lambda_1}(z | x)}\left[\log \frac{p_{\theta}(x | Z) p_{a, \lambda_2}(Z)}{q_{\alpha, \lambda_1}(Z | x)}\right]
\end{align}
where $p_{a, \lambda_2}(z)$ is a Concrete density with location $a$ and temperature $\lambda_2$. At test time we evaluate the discrete lowerbound $\mathcal{L}_1(\theta, a, \alpha)$. Naively implementing Eq. \ref{eq:concrelaxexample2} will result in numerical issues. We discuss this and other details in Appendix \ref{appendix:Concreterelaxations}.

Thus, the basic paradigm we propose is the following: during training replace every discrete node with a Concrete node at some fixed temperature (or with an annealing schedule). The graphs are identical up to the $\softmax$ / $\argmax$ computations, so the parameters of the relaxed graph and discrete graph are the same. When an objective depends on the log-probability of discrete variables in the SCG, as the variational lowerbound does, we propose that the log-probability terms are also ``relaxed'' to represent the true distribution of the relaxed node. At test time the original discrete loss is evaluated. This is possible, because the discretization of any Concrete distribution has a closed form mass function, and the relaxation of any discrete distribution into a Concrete distribution has a closed form density. This is not always possible. For example, the multinomial probit model---the Gumbel-Max trick with Gaussians replacing Gumbels---does not have a closed form mass.

The success of Concrete relaxations will depend on the choice of temperature during training. It is important that the relaxed nodes are not able to represent a precise real valued mode in the interior of the simplex as in Figure \ref{subfig:hightemp}.  If this is the case, it is possible for the relaxed random variable to communicate much more than $\log_2(n)$ bits of information about its $\alpha$ parameters. This might lead the relaxation to prefer the interior of the simplex to the vertices, and as a result there will be a large integrality gap in the overall performance of the discrete graph. 
Therefore Proposition \ref{prop:conc} \ref{itm:convex}  is a conservative guideline for generic $n$-ary Concrete relaxations; at temperatures lower than $(n-1)^{-1}$ we are guaranteed not to have any modes in the interior for any $\alpha \in (0, \infty)^n$. We discuss the subtleties of choosing the temperatures in more detail in Appendix \ref{appendix:Concreterelaxations}. Ultimately the best choice of $\lambda$ and the performance of the relaxation for any specific $n$ will be an empirical question.

%% file: related.tex
\section{Related Work}
\label{sec:related}
Perhaps the most common distribution over the simplex is the Dirichlet with density 
$p_{\alpha}(x) \propto \prod_{k=1}^n x_k^{\alpha_k - 1}$ on $x \in \Delta^{n-1}$. The Dirichlet can be characterized by strong independence properties, and a great deal of work has been done to generalize it \citep{connor1969concepts, aitchison1985general, rayens1994dependence, favaro2011simplex}. Of note is the Logistic Normal distribution \citep{atchison1980logistic}, which can be simulated by taking the softmax of $n-1$ normal random variables and an $n$th logit that is deterministically zero. The Logistic Normal is an important distribution, because it can effectively model correlations within the simplex \citep{blei2006correlated}. To our knowledge the Concrete distribution does not fall completely into any family of distributions previously described. For $\lambda \leq 1$ the Concrete is in a class of normalized infinitely divisible distributions (S. Favaro, personal communication), and the results of \citet{favaro2011simplex} apply.

The idea of using a softmax of Gumbels as a relaxation for a discrete random variable was concurrently considered by \citep{2016arXiv161101144J}, where it was called the Gumbel-Softmax. They do not use the density in the relaxed objective, opting instead to compute all aspects of the graph, including discrete log-probability computations, with the relaxed stochastic state of the graph. In the case of variational inference, this relaxed objective is not a lower bound on the marginal likelihood of the observations, and care needs to be taken when optimizing it.
The idea of using sigmoidal functions with additive input noise to approximate discreteness is also not a new idea. \citep{frey1997continuous} introduced nonlinear
Gaussian units which computed their activation by passing Gaussian noise with
the mean and variance specified by the input to the unit through a
nonlinearity, such as the logistic function.
\citet{salakhutdinov2009semantic} binarized real-valued codes of an autoencoder by adding (Gaussian) noise to the
logits before passing them through the logistic function.
Most recently, to avoid the difficulty
associated with likelihood-ratio methods \citep{kocisky2016parsing} relaxed the discrete
sampling operation by sampling a vector of Gaussians instead and passing those through
a softmax.

There is another family of gradient estimators that have been studied in the context of training neural networks with discrete units. These are usually collected under the umbrella of straight-through estimators \citep{bengio2013estimating, raiko2014techniques}. The basic idea they use is passing forward discrete values, but taking gradients through the expected value. They have good empirical performance, but have not been shown to be the estimators of any loss function. This is in contrast to gradients from Concrete relaxations, which are biased with respect to the discrete graph, but unbiased with respect to the continuous one.

%% file: experiments.tex
\section{Experiments}
\label{sec:experiments}

\subsection{Protocol}

The aim of our experiments was to evaluate the effectiveness of the gradients of Concrete relaxations for optimizing SCGs with discrete nodes. We considered the tasks in \citep{MnihRezende2016}: structured output prediction and density estimation. Both tasks are difficult optimization problems involving fitting probability distributions with hundreds of latent discrete nodes. We compared the performance of Concrete reparameterizations to two state-of-the-art score function estimators: VIMCO \citep{MnihRezende2016} for optimizing the multisample variational objective ($m>1$) and NVIL \citep{MnihGregor2014} for optimizing the single-sample one ($m=1$). We performed the experiments using the MNIST and Omniglot datasets. These are datasets of $28\times28$ images of handwritten digits (MNIST) or letters (Omniglot). For MNIST we used the fixed binarization of \citet{SalMurray08} and the standard 50,000/10,000/10,000 split into training/validation/testing sets. For Omniglot we sampled a fixed binarization and used the standard 24,345/8,070 split into training/testing sets. We report the negative log-likelihood (NLL) of the discrete graph on the test data as the performance metric.

All of our models were neural networks with layers of $n$-ary discrete stochastic nodes with values on the corners of the hypercube $\{-1,1\}^{\log_2(n)}$. The distributions were parameterized by $n$ real values $\log \alpha_k \in \R$, which we took to be the logits of a discrete random variable $D \sim \Discrete(\alpha)$ with $n$ states. Model descriptions are of the form ``(200V--200H$\sim$784V)'', read from left to right. This describes the order of conditional sampling, again from left to right, with each integer representing the number of stochastic units in a layer. The letters V and H represent observed and latent variables, respectively. If the leftmost layer is H, then it was sampled unconditionally from some parameters. Conditioning functions are described by $\{$--, $\sim\}$, where ``--'' means a linear function of the previous layer and ``$\sim$'' means a non-linear function. A ``layer'' of these units is simply the concatenation of some number of independent nodes whose parameters are determined as a function the previous layer. For example a 240 binary layer is a factored distribution over the $\{-1,1\}^{240}$ hypercube. Whereas a 240 $8$-ary layer can be seen as a distribution over the same hypercube where each of the 80 triples of units are sampled independently from an 8 way discrete distribution over $\{-1,1\}^3$. All models were initialized with the heuristic of \citet{glorot2010understanding} and optimized using Adam \citep{kingma2014adam}. All temperatures were fixed throughout training. Appendix \ref{appendix:experimentdetails} for hyperparameter details. 

\subsection{Density Estimation}
\begin{table}[t]
\vspace{-\baselineskip}
\begin{center}
{\footnotesize
\begin{tabular}{@{}l @{}l cc cc cc cc cc cc@{}}
 &     & \multicolumn{4}{c}{MNIST NLL} & \multicolumn{4}{c}{Omniglot NLL} \\
\cmidrule(r){3-6} \cmidrule(l){7-10} 
 \multirow{2}{*}{\shortstack[l]{binary \\ model}} &     & \multicolumn{2}{c}{Test} & \multicolumn{2}{c}{Train} & \multicolumn{2}{c}{Test} & \multicolumn{2}{c}{Train} \\
\cmidrule(r){3-4} \cmidrule(l){5-6} \cmidrule(lr){7-8} \cmidrule(l){9-10} 
& $\iwaess$ & Concrete  &  VIMCO  & Concrete   &  VIMCO  & Concrete  &  VIMCO  & Concrete  & VIMCO  \\ 
\toprule
\multirow{3}{*}{\shortstack[l]{(200H \\-- 784V)}}\hspace{0.5em}
             & 1    & 107.3 & {\bf104.4} & 107.5  & \bf104.2 & 118.7 & \bf115.7 &117.0 &  \bf112.2 \\
             & 5    & 104.9 & \bf 101.9 & 104.9  & \bf 101.5 & 118.0 & \bf113.5 &115.8 & \bf 110.8 \\
             & 50   & 104.3 &  \bf 98.8 & 104.2  &  \bf  98.3 & 118.9 &\bf 113.0 &115.8 & \bf 110.0 \\
              \midrule
\multirow{3}{*}{\shortstack[l]{(200H\\-- 200H\\-- 784V)}}
             & 1    & 102.1 &\bf 92.9 &102.3  &   \bf 91.7 & 116.3 & \bf109.2 &114.4 & \bf 104.8 \\
             & 5    &  99.9 & \bf91.7 &  100.0  &\bf  90.8 & 116.0 & \bf107.5 & 113.5 & \bf103.6 \\
             & 50   &  99.5 & \bf 90.7 &  99.4  &\bf  89.7 & 117.0 & \bf108.1 &113.9 & \bf 103.6 \\
              \midrule
\multirow{3}{*}{\shortstack[l]{(200H\\$\sim$784V)}}
             & 1    &\bf  92.1 &  93.8 & \bf 91.2  &   91.5 &\bf 108.4 & 116.4 &\bf103.6 &  110.3 \\
             & 5    & \bf 89.5 &   91.4 &\bf 88.1  &   88.6 & \bf107.5 &118.2 &\bf 101.4 &  102.3 \\
             & 50   &\bf  88.5 &   89.3 &\bf 86.4  &  86.5 &\bf 108.1 & 116.0 &\bf100.5 &  100.8 \\
              \midrule
\multirow{3}{*}{\shortstack[l]{(200H\\$\sim$200H\\$\sim$784V)}}
             & 1    & \bf 87.9 & 88.4 & 86.5  &  \bf  85.8 &\bf 105.9 &111.7 & \bf 100.2 &  105.7 \\
             & 5    & \bf 86.3 &  86.4 & 84.1  & \bf   82.5 &\bf 105.8 &108.2 &  \bf 98.6 & 101.1 \\
             & 50   &  85.7 & \bf 85.5 & 83.1  & \bf  81.8 &\bf 106.8 &113.2 &  97.5 & \bf  95.2 \\
              \bottomrule
\end{tabular}}
\end{center}
\caption{Density estimation with binary latent variables. When $m=1$, VIMCO stands for NVIL.}
\label{tbl:vimcocompare}
\vspace{-\baselineskip}
\end{table}

Density estimation, or generative modelling, is the problem of fitting the distribution of data. We took the
latent variable approach described in \sect{sec:lvm_training} and
trained the models by optimizing the variational objective
$\mathcal{L}_m(\theta, \phi)$ given by \eqn{eqn:iw_objective} averaged uniformly over minibatches of data points $x$. Both our
generative models $p_{\theta}(z, \  x)$ and variational distributions
$q_{\phi}(z \given x)$ were parameterized with neural networks as described above.
We trained models with $\mathcal{L}_m(\theta, \phi)$ for $m \in \{1, 5, 50\}$ and approximated the NLL with $\mathcal{L}_{50,000}(\theta, \phi)$ averaged uniformly over the whole dataset. 

The results are shown in \tbl{tbl:vimcocompare}. In general, VIMCO outperformed Concrete relaxations for linear models and Concrete relaxations outperformed VIMCO for non-linear models. We also tested the effectiveness of Concrete relaxations on generative models with $n$-ary layers on the $\mathcal{L}_5(\theta, \phi)$ objective. The best $4$-ary model achieved test/train NLL 86.7/83.3, the best 8-ary achieved 87.4/84.6 with Concrete relaxations, more complete results in Appendix \ref{appendix:extraresults}. The relatively poor performance of the 8-ary model may be because moving from 4 to 8 results in a more difficult objective without much added capacity. As a control we trained $n$-ary models using logistic normals as relaxations of discrete distributions (with retuned temperature hyperparameters). Because the discrete zero temperature limit of logistic Normals is a multinomial probit whose mass function is not known, we evaluated the discrete model by sampling from the discrete distribution parameterized by the logits learned during training. The best 4-ary model achieved test/train NLL  of 88.7/85.0, the best 8-ary model achieved 89.1/85.1.

\subsection{Structured Output Prediction}

\begin{figure}[t]
\vspace{-\baselineskip}
\begin{subfigure}{0.55\textwidth}
\begin{center}
{\footnotesize
\begin{tabular}{@{}l @{}l @{}c @{}c @{}c @{}c@{}}
\multirow{2}{*}{\shortstack[l]{binary \\ model}}  & & \multicolumn{2}{c}{Test NLL} &  \multicolumn{2}{c}{Train NLL} \\
\cmidrule(r){3-4} \cmidrule(l){5-6} 
 & $m$\hspace{0.5em}    & Concrete\hspace{0.5em} & VIMCO\hspace{0.5em} & Concrete\hspace{0.5em} & VIMCO\\
\toprule
\multirow{3}{*}{\shortstack[l]{(392V--240H\\--240H--392V)}}\hspace{0.5em}   &  1 & \bf 58.5  & 61.4 & \bf 54.2 &  59.3 \\
                                            &  5 &   \bf54.3  & 54.5 & \bf 49.2 &  52.7 \\
                                            & 50 &  53.4  &\bf 51.8 & \bf 48.2 &  49.6 \\
\midrule
\multirow{3}{*}{\shortstack[l]{(392V--240H\\--240H--240H\\--392V)}}  &  1 &\bf 56.3 &59.7 & \bf 51.6 &  58.4 \\
                                                  &  5 &\bf 52.7 & 53.5 &\bf46.9 &  51.6 \\
                                                  & 50 & 52.0 & \bf 50.2 &\bf 45.9 &  47.9 \\
\bottomrule
\end{tabular}}
\end{center}
\label{tbl:structpred}
\end{subfigure}
\begin{subfigure}{0.44\textwidth}
\begin{center}
          \resizebox{\linewidth}{!}{\includegraphics{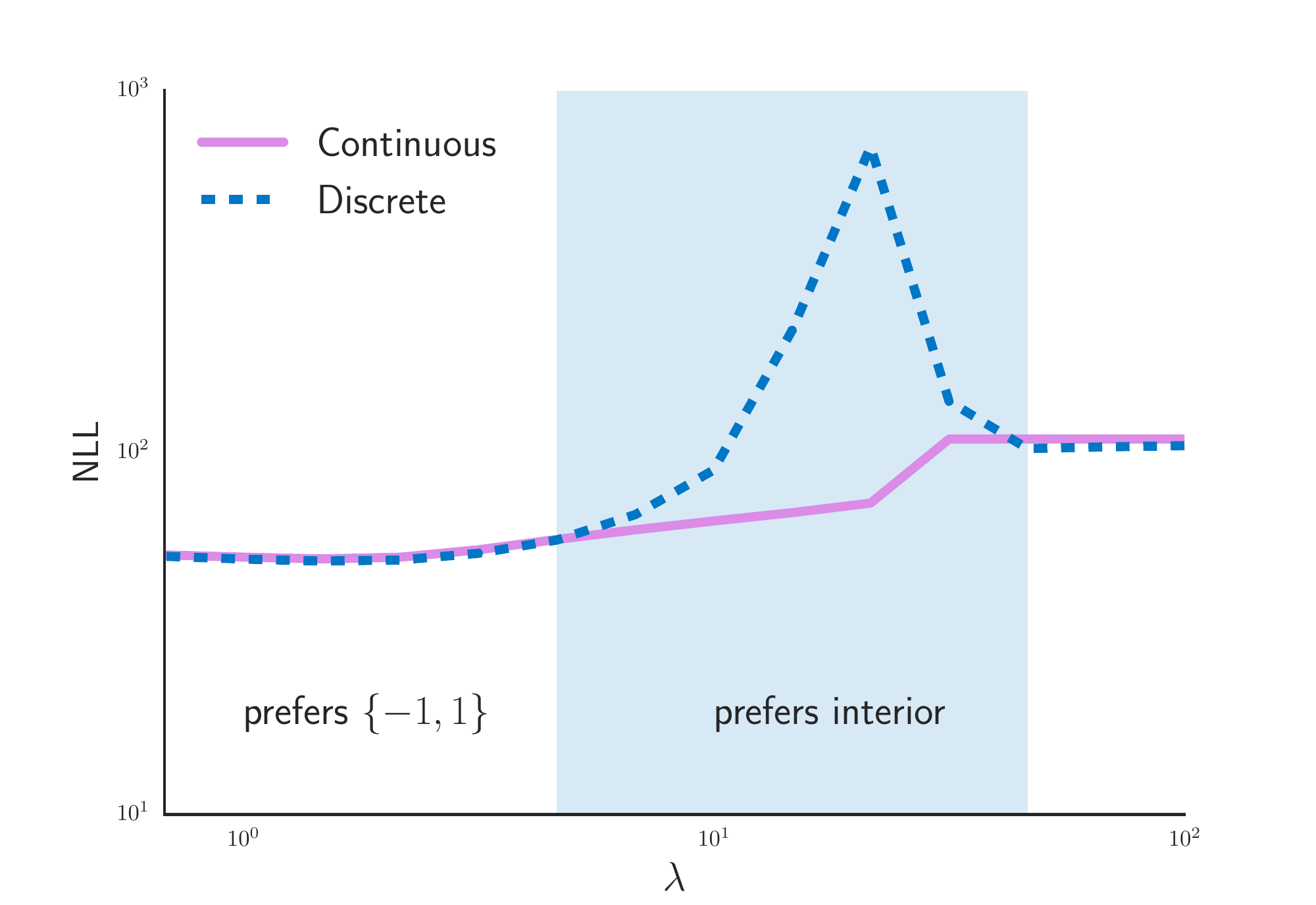}}
\end{center}
\end{subfigure}
\caption{Results for structured prediction on MNIST comparing Concrete relaxations to VIMCO. When $m=1$ VIMCO stands for NVIL. The plot on the right shows the objective (lower is better) for the continuous and discrete graph trained at temperatures $\lambda$. In the shaded region, units prefer to communicate real values in the interior of $(-1,1)$ and the discretization suffers an integrality gap.}
\label{tbl:structpred}
\vspace{-\baselineskip}
\end{figure}

Structured output prediction is concerned with modelling the high-dimensional
distribution of the observation given a context and can be seen as conditional
density estimation. We considered the task of predicting the bottom half
$x_1$ of an image of an MNIST digit given its top half $x_2$, as introduced by
\citet{raiko2014techniques}. We followed
\citet{raiko2014techniques} in using a model with layers of discrete stochastic
units between the context and the observation. Conditioned on the top half
$x_2$ the network samples from a distribution $p_{\phi}(z \given x_2)$ over
layers of stochastic units $z$ then predicts $x_1$ by sampling from a
distribution $p_{\theta}(x_1 \given z)$. The training objective for a single pair $(x_1,
x_2)$ is
\begin{align*}
\mathcal{L}^{SP}_m(\theta, \phi) = \expect_{Z_i \sim p_{\phi}(z | x_2) }\left[ \log\left( \frac{1}{m} \sum_{i=1}^m p_{\theta}(x_1 \given Z_i) \right)\right].
\end{align*}
This objective is a special case of $\mathcal{L}_m(\theta, \phi)$
(\eqn{eqn:iw_objective}) where we use the prior $p_{\phi}(z | x_2)$ as the
variational distribution. Thus, the objective is a lower
bound on $\log p_{\theta, \phi}(x_1 \given x_2)$.

We trained the models by optimizing $\mathcal{L}^{SP}_m(\theta, \phi)$ for $m\in \{1,5,50\}$ averaged uniformly over minibatches and evaluated them by computing $\mathcal{L}^{SP}_{100}(\theta, \phi)$ averaged uniformly over the entire dataset.
The results are shown in Figure \ref{tbl:structpred}. Concrete relaxations more uniformly outperformed VIMCO in this instance. We also trained $n$-ary (392V--240H--240H--240H--392V) models on the  $\mathcal{L}^{SP}_1(\theta, \phi)$ objective using the best temperature hyperparameters from density estimation. 4-ary achieved a test/train NLL of 55.4/46.0 and 8-ary achieved 54.7/44.8. As opposed to density estimation, increasing arity uniformly improved the models.
We also investigated the hypothesis that for higher temperatures Concrete relaxations might prefer the interior of the interval to the boundary points $\{-1,1\}$. Figure \ref{tbl:structpred} was generated with binary  (392V--240H--240H--240H--392V) model trained on  $\mathcal{L}^{SP}_1(\theta, \phi)$.

%% file: conclusion.tex
\section{Conclusion}
\label{sec:conclusion}

We introduced the Concrete distribution, a continuous relaxation of discrete random variables. The Concrete distribution is a new distribution on the simplex with a closed form density parameterized by a vector of positive location parameters and a positive temperature. Crucially, the zero temperature limit of every Concrete distribution corresponds to a discrete distribution, and any discrete distribution can be seen as the discretization of a Concrete one. The application we considered was training stochastic computation graphs with discrete stochastic nodes. The gradients of Concrete relaxations are biased with respect to the original discrete objective, but they are low variance unbiased estimators of a continuous surrogate objective. We showed in a series of experiments that stochastic nodes with Concrete distributions can be used effectively to optimize the parameters of a stochastic computation graph with discrete stochastic nodes.
We did not find that annealing or automatically tuning the temperature was important for these experiments, but it remains interesting and possibly valuable future work.

%% file: appendix.tex
\section{Proof of Proposition \ref{prop:conc}}
\label{appendix:prop1}

Let $X  \sim \Concrete(\alpha, \lambda)$ with location parameters $\alpha \in (0, \infty)^n$ and temperature $\lambda \in (0, \infty)$.

\begin{enumerate}
\item Let $G_k \sim \Gumbel$ i.i.d., consider
\begin{align*}
Y_k = \frac{\exp((\log\alpha_k + G_k)/\lambda)}{\sum_{i=1}^{n} \exp((\log\alpha_i + G_i)/\lambda)}
\end{align*}
Let $Z_k = \log \alpha_k + G_k$, which has density
\begin{align*}
\alpha_k \exp(-z_k) \exp(- \alpha_k\exp(-z_k))
\end{align*}
We will consider the invertible transformation
\begin{align*}
F(z_1, \ldots, z_{n}) &= (y_1, \ldots, y_{n-1}, c)\\
\intertext{where}
y_k &= \exp(z_k/\lambda) c^{-1}\\
c &= \sum_{i=1}^{n} \exp(z_i/\lambda)\\
\end{align*}
then
\begin{align*}
F^{-1}(y_1, \ldots, y_{n-1}, c) = \left(\lambda(\log y_1 + \log c), \ldots,\lambda(\log y_{n-1} + \log c),  \lambda(\log y_n + \log c)\right)
\end{align*}
where $y_{n} = 1 - \sum_{i=1}^{n-1} y_i$.
This has Jacobian
\begin{align*}
\left[\begin{array}{ccccccc}
\lambda y_1^{-1} & 0 & 0 &0&  \ldots & 0&\lambda c^{-1}\\
0 & \lambda y_2^{-1} & 0 & 0 & \ldots & 0& \lambda c^{-1}\\
0 & 0 & \lambda y_3^{-1} & 0 & \ldots & 0&\lambda c^{-1}\\
 &  &  \vdots & & & & \\
-\lambda y_{n}^{-1} & -\lambda y_{n}^{-1} & -\lambda y_{n}^{-1} & -\lambda y_n^{-1} & \ldots & -\lambda y_{n}^{-1} & \lambda c^{-1}
\end{array}\right]
\end{align*}
by adding $ y_i / y_{n}$ times each of the top $n-1$ rows to the bottom row we see that this Jacobian has the same determinant as
\begin{align*}
\left[\begin{array}{ccccccc}
\lambda y_1^{-1} & 0 & 0 &0&  \ldots & 0&\lambda c^{-1}\\
0 &  \lambda y_2^{-1} & 0 & 0 & \ldots & 0&\lambda c^{-1}\\
0 & 0 & \lambda y_3^{-1} & 0 & \ldots & 0&\lambda c^{-1}\\
 &  &  \vdots & & & & \\
0 &0 & 0 & 0 & \ldots & 0&\lambda (c y_{n})^{-1}
\end{array}\right]
\end{align*}
and thus the determinant is equal to 
\begin{align*}
\frac{\lambda^n}{c\prod_{i=1}^{k} y_i}
\end{align*}
all together we have the density
\begin{align*}
\frac{\lambda^{n}\prod_{k=1}^{n}\alpha_k\exp(-\lambda\log y_k - \lambda \log c) \exp(- \alpha_k\exp(-\lambda\log y_k - \lambda \log c))}{c\prod_{i=1}^{n} y_i}
\end{align*}
with $r = \log c$ change of variables we have density
\begin{align*}
&\frac{\lambda^{n}\prod_{k=1}^{n}\alpha_k\exp(- \lambda r) \exp(- \alpha_k\exp(-\lambda\log y_k - \lambda r))}{\prod_{i=1}^{n} y_i^{\lambda + 1}} = \\
&\frac{\lambda^n \prod_{k=1}^n \alpha_k}{\prod_{i=1}^{n} y_i^{\lambda + 1}}\exp(- n\lambda r) \exp(- \sum_{i=1}^{n} \alpha_i\exp(-\lambda \log y_i - \lambda r )) = \\
\intertext{letting $\gamma = \log (\sum_{n=1}^{n} \alpha_k y_k^{-\lambda})$}
&\frac{\lambda^n \prod_{k=1}^n \alpha_k}{\prod_{i=1}^{n} y_i^{\lambda + 1}\exp(\gamma)}\exp(- n\lambda r + \gamma) \exp(- \exp(- \lambda r + \gamma)) = \\
\intertext{integrating out $r$}
&\frac{\lambda^n \prod_{k=1}^n \alpha_k}{\prod_{i=1}^{n} y_i^{\lambda + 1}\exp(\gamma)}\left(\frac{\exp(- \gamma n + \gamma) \Gamma(n)}{\lambda}\right) = \\
&\frac{\lambda^{n-1} \prod_{k=1}^n \alpha_k}{\prod_{i=1}^{n} y_i^{\lambda + 1}}\left(\exp(- \gamma n) \Gamma(n)\right) = \\
& (n-1)! \lambda^{n-1} \frac{\prod_{k=1}^n \alpha_k y_k^{-\lambda - 1}}{(\sum_{n=1}^{n} \alpha_k y_k^{-\lambda})^n}
\end{align*}
Thus $Y \overset{d}{=} X$.
\item Follows directly from \ref{itm:reparam} and the Gumbel-Max trick \citep{maddison2016poisson}.
\item Follows directly from \ref{itm:reparam} and the Gumbel-Max trick \citep{maddison2016poisson}.
\item Let $\lambda \leq (n-1)^{-1}$. The density of $X$ can be rewritten as
\begin{align*}
p_{\alpha, \lambda}(x) &\propto \prod_{k=1}^n \frac{\alpha_k y^{-\lambda - 1}}{\sum_{i=1}^n \alpha_i y_i^{-\lambda}}\\
&=\prod_{k=1}^n \frac{\alpha_k y_k^{\lambda(n-1) - 1}}{\sum_{i=1}^n \alpha_i \prod_{j \neq i} y_j^{\lambda}}
\end{align*}
Thus, the log density is up to an additive constant $C$
\begin{align*}
\log p_{\alpha, \lambda}(x) = \sum_{k=1}^n (\lambda (n-1) - 1)\log y_k - n \log\left(\sum_{k=1}^n \alpha_k \prod_{j\neq k} y_j^{\lambda} \right) + C
\end{align*}
If $\lambda \leq (n - 1)^{-1}$, then the first $n$ terms are convex, because $-\log$ is convex. For the last term, $-\log$ is convex and non-increasing and $\prod_{j \neq k} y_j^{\lambda}$ is concave for $\lambda \leq (n-1)^{-1}$. Thus, their composition is convex. The sum of convex terms is convex, finishing the proof.
\end{enumerate}

\section{The Binary Special Case}
\label{appendix:bingumbelmax}

Bernoulli random variables are an important special case of discrete distributions taking states in $\{0, 1\}$. Here we consider the binary special case of the Gumbel-Max trick from Figure \ref{subfig:discrete} along with the corresponding Concrete relaxation.
 
Let $D \sim \Discrete(\alpha)$ for $\alpha \in (0, \infty)^2$ be a two state discrete random variable on $\{0,1\}^2$ such that $D_1 + D_2 = 1$, parameterized as in Figure \ref{subfig:discrete} by $\alpha_1, \alpha_2 > 0$:
\begin{align}
\proba(D_1 = 1) = \frac{\alpha_1}{\alpha_1 + \alpha_2}
\end{align}
The distribution is degenerate, because $D_1 = 1-D_2$. Therefore we consider just $D_1$. Under the Gumbel-Max reparameterization, the event that $D_1 = 1$ is the event that $\{G_1 + \log \alpha_1 > G_2 + \log \alpha_2\}$ where $G_k \sim \Gumbel$ i.i.d. The difference of two Gumbels is a Logistic distribution $G_1 - G_2 \sim \Logistic$, which can be sampled in the following way, $G_1 - G_2 \overset{d}{=} \log U - \log(1-U)$ where $U\sim\Uniform(0,1)$. So, if $\alpha = \alpha_1 / \alpha_2$, then we have
\begin{align}
\proba(D_1 = 1) = \proba(G_1 + \log \alpha_1 > G_2 + \log \alpha_2) = \proba(\log U - \log(1-U) + \log \alpha > 0)
\end{align}
Thus, $D_1 \overset{d}{=} H(\log \alpha + \log U - \log(1-U))$, where $H$ is the unit step function.

Correspondingly, we can consider the Binary Concrete relaxation that results from this process. As in the $n$-ary case, we consider the sampling routine for a Binary Concrete random variable $X \in (0,1)$ first. To sample $X$, sample $L \sim \Logistic$ and set
\begin{align}
X = \frac{1}{1 + \exp(- (\log \alpha + L) / \lambda)}
\end{align}
We define the Binary Concrete random variable $X$ by its density on the unit interval.
\begin{defn}[Binary Concrete Random Variables]
\label{def:binconcrete}
Let $\alpha \in (0, \infty)$ and $\lambda \in (0, \infty)$. $X \in (0, 1)$ has a Binary Concrete distribution $X \sim \BinaryConcrete(\alpha, \lambda)$ with location $\alpha$ and temperature $\lambda$, if its density is:
\begin{align}
p_{\alpha, \lambda}(x) =  \frac{\lambda \alpha x^{-\lambda - 1}(1-x)^{-\lambda - 1}}{(\alpha x^{-\lambda} + (1-x)^{-\lambda})^2}.
\end{align}
\end{defn}
We state without proof the special case of Proposition \ref{prop:conc} for Binary Concrete distributions
\begin{prop}[Some Properties of Binary Concrete Random Variables]
Let $X  \sim \BinaryConcrete(\alpha, \lambda)$ with location parameter $\alpha \in (0, \infty)$ and temperature $\lambda \in (0, \infty)$, then
\begin{enumerate}[label=(\alph*)]
\item (Reparameterization) If $L \sim \Logistic$, then $X \overset{d}{=} \frac{1}{1 + \exp(- (\log \alpha + L) / \lambda)}$,
\item (Rounding) $\proba\left(X > 0.5 \right) = \alpha / (1 + \alpha)$,
\item (Zero temperature) $\proba\left(\lim_{\lambda \to 0} X = 1\right) = \alpha / (1 + \alpha)$,
\item (Convex eventually) If $\lambda \leq 1$, then $p_{\alpha, \lambda}(x)$ is log-convex in $x$.
\end{enumerate}
\end{prop}

We can generalize the binary circuit beyond Logistic random variables. Consider an arbitrary random variable $X$ with infinite support on $\R$. If $\Phi : \R \to [0,1]$ is the CDF of $X$, then
\begin{align*}
\proba(H(X) = 1) = 1 - \Phi(0)
\end{align*}
If we want this to have a Bernoulli distribution with probability $\alpha/(1+\alpha)$, then we should solve the equation
\begin{align*}
1 - \Phi(0) = \frac{\alpha}{1+\alpha}.
\end{align*}
This gives $\Phi(0) = 1/(1+\alpha)$, which can be accomplished by relocating the random variable $Y$ with CDF $\Phi$ to be $X = Y - \Phi^{-1}(1/(1+\alpha))$.
 
\section{Using Concrete Relaxations}
\label{appendix:Concreterelaxations}

In this section we include some tips for implementing and using the Concrete distribution as a relaxation. We use the following notation
\begin{align*}\sigma(x) &= \frac{1}{1 + \exp(-x)} \;\;\;\;\;\;\;\; \logsumexp_{k=1}^n \{x_k\} = \log \left(\sum_{k=1}^n \exp(x_k)\right)\end{align*}
Both sigmoid and log-sum-exp are common operations in libraries like TensorFlow or theano. 

\subsection{The Basic Problem}

For the sake of exposition, we consider a simple variational autoencoder with a single discrete random variable and objective $\mathcal{L}_1(\theta, a, \alpha)$ given by \eqn{eqn:iw_objective} for a single data point $x$. This scenario will allow us to discuss all of the decisions one might make when using Concrete relaxations.

In particular, let $P_{a}(d)$ be the mass function of some $n$-dimensional one-hot discrete $D \sim \Discrete(a)$ with $a \in (0, \infty)^n$, let $p_{\theta}(x | d)$ be some likelihood (possibly computed by a neural network), which is a continuous function of $d$ and parameters $\theta$, let $D \sim \Discrete(\alpha(x))$ be a one-hot discrete random variable in $(0, 1)^n$ whose unnormalized probabilities $\alpha(x) \in (0, \infty)^n$ are some function (possible a neural net with its own parameters) of $x$. Let $Q_{\alpha}(d | x)$ be the mass function of $D$. Then, we care about optimizing
\begin{align}
\label{appendix:concrelaxeq}
\mathcal{L}_1(\theta, a, \alpha) = \expect_{D \sim Q_{\alpha}(d | x)}\left[\log \frac{p_{\theta}(x | D) P_{a}(D)}{Q_{\alpha}(D | x)}\right]
\end{align}
with respect to $\theta$, $a$, and any parameters in $\alpha$ from samples of the SCG required to simulate an estimator of $\mathcal{L}_1(\theta, a, \alpha)$.

\subsection{What you might relax and why}

The first consideration when relaxing an estimator of Eq. \ref{appendix:concrelaxeq} is how to relax the stochastic computation. The only sampling required to simulate $\mathcal{L}_1(\theta, a, \alpha)$ is $D \sim \Discrete(\alpha(x))$. The corresponding Concrete relaxation is to sample $Z \sim \Concrete(\alpha(x), \lambda_1)$ with temperature $\lambda_1$ and location parameters are the the unnormalized probabilities $\alpha(x)$ of $D$. Let density $q_{\alpha, \lambda_1}(z | x)$ be the density of $Z$. We get a relaxed objective of the form:
\begin{align}
\label{appendix:eq2}
\expect_{D \sim Q_{\alpha}(d | x)}\left[ \ \cdot \ \right] \ \to \ \expect_{Z \sim q_{\alpha, \lambda_1}(z | x)}\left[ \ \cdot \ \right]
\end{align}
This choice allows us to take derivatives through the stochastic computaitons of the graph.

The second consideration is which objective to put in place of $[\ \cdot \ ]$ in Eq. \ref{appendix:eq2}. We will consider the ideal scenario irrespective of numerical issues. In Subsection \ref{appendix:stochasticnodesub} we address those numerical issues.  The central question is how to treat the expectation of the ratio $P_{a}(D) / Q_{\alpha}(D | x)$ (which is the KL component of the loss) when $Z$ replaces $D$.

There are at least three options for how to modify the objective. They are, (\ref{eq:relaxall}) replace the discrete mass with Concrete densities, (\ref{eq:relaxdiscmass}) relax the computation of the discrete log mass, (\ref{eq:analyticdisckl}) replace it with the analytic discrete KL.
\begin{align}
&\expect_{Z \sim q_{\alpha, \lambda_1}(z | x)}\left[ \log p_{\theta}(x | Z) + \log \frac{p_{a, \lambda_2}(Z)}{q_{\alpha, \lambda_1}(Z | x)}\right] \label{eq:relaxall}\\
&\expect_{Z \sim q_{\alpha, \lambda_1}(z | x)}\left[ \log p_{\theta}(x | Z) + \sum_{i=1}^n Z_i \log\frac{P_{a}(d^{(i)})}{Q_{\alpha}(d^{(i)} | x)}\right] \label{eq:relaxdiscmass}\\
&\expect_{Z \sim q_{\alpha, \lambda_1}(z | x)}\left[ \log p_{\theta}(x | Z) \right] + \sum_{i=1}^n Q_{\alpha}(d^{(i)} | x) \log\frac{P_{a}(d^{(i)})}{Q_{\alpha}(d^{(i)} | x)} \label{eq:analyticdisckl}
\end{align}
where $d^{(i)}$ is a one-hot binary vector with $d_i^{(i)} =1$ and $p_{a, \lambda_2}(z)$ is the density of some Concrete random variable with temperature $\lambda_2$ with location parameters $a$. Although (\ref{eq:analyticdisckl}) or (\ref{eq:relaxdiscmass}) is tempting, we emphasize that these are NOT necessarily lower bounds on $\log p(x)$ in the relaxed model. (\ref{eq:relaxall}) is the only objective guaranteed to be a lower bound:
\begin{align}
\expect_{Z \sim q_{\alpha, \lambda_1}(z | x)}\left[ \log p_{\theta}(x | Z) + \log \frac{p_{a, \lambda_2}(Z)}{q_{\alpha, \lambda_1}(Z | x)}\right]  \leq \log \int p_{\theta}(x | z) p_{a, \lambda_2}(z) \ dx.
\end{align}
For this reason we consider objectives of the form (\ref{eq:relaxall}). Choosing (\ref{eq:analyticdisckl}) or (\ref{eq:relaxdiscmass}) is possible, but the value of these objectives is not interpretable and one should early stop otherwise it will overfit to the spurious ``KL'' component of the loss. We now consider practical issues with (\ref{eq:relaxall}) and how to address them. All together we can interpret $q_{\alpha, \lambda_1}(z | x)$ as the Concrete relaxation of the variational posterior and $p_{a, \lambda_2}(z)$ the relaxation of the prior.

\subsection{Which random variable to treat as the stochastic node}
\label{appendix:stochasticnodesub}

When implementing a SCG like the variational autoencoder example, we need to compute log-probabilities of Concrete random variables. This computation can suffer from underflow, so where possible it's better to take a different node on the relaxed graph as the stochastic node on which log-likelihood terms are computed. For example, it's tempting in the case of Concrete random variables to treat the Gumbels as the stochastic node on which the log-likelihood terms are evaluated and the softmax as downstream computation. This will be a looser bound in the context of variational inference than the corresponding bound when treating the Concrete relaxed states as the node.

The solution we found to work well was to work with Concrete random variables in log-space. Consider the following vector in $\R^n$ for location parameters $\alpha \in (0, \infty)^n$ and $\lambda \in (0, \infty)$ and $G_k \sim \Gumbel$,
\begin{align*}
Y_k = \frac{\log \alpha_k + G_k}{\lambda} - \logsumexp_{i=1}^n\left\{\frac{\log \alpha_i + G_i}{\lambda} \right\}
\end{align*}
$Y \in \R^n$ has the property that $\exp(Y) \sim \Concrete(\alpha, \lambda)$, therefore we call $Y$ an $\ExpConcrete(\alpha, \lambda)$. The advantage of this reparameterization is that the KL terms of a variational loss are invariant under \emph{invertible} transformation. $\exp$ is invertible, so the KL between two ExpConcrete random variables is the same as the KL between two Concrete random variables. The log-density $\log \kappa_{\alpha, \lambda} (y)$ of an $\ExpConcrete(\alpha, \lambda)$ is also simple to compute:
\begin{align*}
\log \kappa_{\alpha, \lambda} (y) = \log((n-1)!) + (n-1)\log \lambda + \left(\sum_{k=1}^n \log \alpha_k - \lambda y_k \right) -  n\logsumexp_{k=1}^n\left\{\log \alpha_k -\lambda y_k\right\}
\end{align*}
for $y \in \R^n$ such that $\logsumexp_{k=1}^n\{y_k\} = 0$. Note that the sample space of the ExpConcrete distribution is still interpretable in the zero temperature limit. In the limit of $\lambda \to 0$ $\ExpConcrete$ random variables become discrete random variables over the one-hot vectors of $d \in \{-\infty, 0\}^n$ where $\logsumexp_{k=1}^n \{d_k\} = 0$. $\exp(Y)$ in this case results in the one-hot vectors in $\{0,1\}^n$.

\subsubsection{$n$-ary Concrete}

Returning to our initial task of relaxing $\mathcal{L}_1(\theta, a, \alpha)$, let $Y \sim \ExpConcrete(\alpha(x), \lambda_1)$  with density $\kappa_{\alpha, \lambda_1}(y | x)$ be the ExpConcrete latent variable corresponding to the Concrete relaxation $q_{\alpha, \lambda_1}(z | x)$ of the variational posterior $Q_{\alpha}(d | x)$. Let $\rho_{a, \lambda_1}(y)$ be the density of an ExpConcrete random variable corresponding to the Concrete relaxation $p_{a, \lambda_2}(z)$ of $P_{a}(d)$. All together we can see that
\begin{align}
\expect_{Z \sim q_{\alpha, \lambda_1}(z | x)}\left[ \log p_{\theta}(x | Z) + \log \frac{p_{a, \lambda_2}(Z)}{q_{\alpha, \lambda_1}(Z | x)}\right]  = \expect_{Y \sim {\kappa}_{\alpha, \lambda_1}(y | x)}\left[ \log p_{\theta}(x | \exp(Y)) + \log \frac{\rho_{a, \lambda_2}(Y)}{\kappa_{\alpha, \lambda_1}(Y | x)}\right] 
\end{align}
Therefore, we used $\ExpConcrete$ random variables as the stochastic nodes and treated $\exp$ as a downstream computation. The relaxation is then,
\begin{align}
\mathcal{L}_1(\theta, a, \alpha) \overset{\text{relax}}{\leadsto} \expect_{Y \sim {\kappa}_{\alpha, \lambda_1}(y | x)}\left[ \log p_{\theta}(x | \exp(Y)) + \log \frac{\rho_{a, \lambda_2}(Y)}{\kappa_{\alpha, \lambda_1}(Y | x)}\right] ,
\end{align}
and the objective on the RHS is fully reparameterizable and what we chose to optimize.

\subsubsection{Binary Concrete}

In the binary case, the logistic function is invertible, so it makes most sense to treat the logit plus noise as the stochastic node. In particular, the binary random node was sample from:
\begin{align}
\label{eq:ourlogistic}
Y = \frac{\log \alpha + \log U - \log (1-U)}{\lambda}
\end{align}
where $U \sim \Uniform(0,1)$ and always followed by $\sigma$ as downstream computation. $\log U - \log (1-U)$ is a Logistic random variable, details in the cheat sheet, and so the log-density $\log g_{\alpha, \lambda}(y)$ of this node (before applying $\sigma$) is
\begin{align*}
\log g_{\alpha, \lambda}(y)= \log\lambda  -\lambda y + \log \alpha - 2\log(1+ \exp(-\lambda y + \log \alpha))
\end{align*}
All together the relaxation in the binary special case would be
\begin{align}
\mathcal{L}_1(\theta, a, \alpha) \overset{\text{relax}}{\leadsto} \expect_{Y \sim {g}_{\alpha, \lambda_1}(y | x)}\left[ \log p_{\theta}(x | \sigma(Y)) + \log \frac{f_{a, \lambda_2}(Y)}{g_{\alpha, \lambda_1}(Y | x)}\right] ,
\end{align}
where $f_{a, \lambda_2}(y)$ is the density of a Logistic random variable sampled via Eq. \ref{eq:ourlogistic} with location $a$ and temperature $\lambda_2$.

This section had a dense array of densities, so we summarize the relevant ones, along with how to sample from them, in Appendix \ref{appendix:cheatsheet}.

\subsection{Choosing the temperature}
The success of Concrete relaxations will depend heavily on the choice of temperature during training. It is important that the relaxed nodes are not able to represent a precise real valued mode in the interior of the simplex as in Figure \ref{subfig:hightemp}. For example, choosing additive Gaussian noise $\epsilon \sim \Normal(0,1)$ with the logistic function $\sigma(x)$ to get relaxed Bernoullis of the form $\sigma(\epsilon + \mu)$ will result in a large mode in the centre of the interval. This is because the tails of the Gaussian distribution drop off much faster than the rate at which $\sigma$ squashes. Even including a temperature parameter does not completely solve this problem; the density of $\sigma((\epsilon + \mu)/ \lambda)$ at \emph{any} temperature still goes to 0 as its approaches the boundaries 0 and 1 of the unit interval. Therefore \ref{itm:convex} of Proposition \ref{prop:conc} is a conservative guideline for generic $n$-ary Concrete relaxations; at temperatures lower than $(n-1)^{-1}$ we are guaranteed not to have any modes in the interior for any $\alpha \in (0, \infty)^n$. In the case of the Binary Concrete distribution, the tails of the Logistic additive noise are balanced with the logistic squashing function and for temperatures $\lambda \leq 1$ the density of the Binary Concrete distribution is log-convex for all parameters $\alpha$, see Figure \ref{subfig:binconvex}. Still, practice will often disagree with theory here. The peakiness of the Concrete distribution increases with $n$, so much higher temperatures are tolerated (usually necessary).

For $n =1$ temperatures $\lambda \leq (n-1)^{-1}$ is a good guideline. For $n > 1$ taking $\lambda \leq (n-1)^{-1}$ is not necessarily a good guideline, although it will depend on $n$ and the specific application. As $n \to \infty$ the Concrete distribution becomes peakier, because the random normalizing constant $\sum_{k=1}^n \exp((\log\alpha_k + G_k)/\lambda)$ grows. This means that practically speaking the optimization can tolerate much higher temperatures than $(n-1)^{-1}$. We found in the cases $n=4$ that $\lambda = 1$ was the best temperature and in $n=8$, $\lambda =2/3$ was the best. Yet $\lambda = 2/3$ was the best single performing temperature across the $n \in \{2,4,8\}$ cases that we considered. We recommend starting in that ball-park and exploring for any specific application.

When the loss depends on a KL divergence between two Concrete nodes, it's possible to give the nodes distinct temperatures. We found this to improve results quite dramatically. In the context of our original problem and it's relaxation: 
\begin{align}
\label{eq:tempchoose}
\mathcal{L}_1(\theta, a, \alpha) \overset{\text{relax}}{\leadsto} \expect_{Y \sim {\kappa}_{\alpha, \lambda_1}(y | x)}\left[ \log p_{\theta}(x | \exp(Y)) + \log \frac{\rho_{a, \lambda_2}(Y)}{\kappa_{\alpha, \lambda_1}(Y | x)}\right] ,
\end{align}
Both $\lambda_1$ for the posterior temperature and $\lambda_2$ for the prior temperature are tunable hyperparameters.

 \section{Experimental Details}
\label{appendix:experimentdetails}

The basic model architectures we considered are exactly analogous to those in \citet{burda2015importance} with Concrete/discrete random variables replacing Gaussians.

\subsection{--- vs $\sim$} The conditioning functions we used were either linear or non-linear. Non-linear consisted of two $\tanh$ layers of the same size as the preceding stochastic layer in the computation graph.

\subsection{$n$-ary layers}
All our models are neural networks with layers of $n$-ary discrete stochastic nodes with $\log_2(n)$-dimensional states on the corners of the hypercube $\{-1,1\}^{\log_2(n)}$. For a generic $n$-ary node sampling proceeds as follows. Sample a $n$-ary discrete random variable $D \sim \Discrete(\alpha)$ for $\alpha \in (0, \infty)^n$.
If $C$ is the $\log_2(n) \times n$ matrix, which lists the corners of the hypercube $\{-1,1\}^{\log_2(n)}$ as columns, then we took $Y = C D$ as downstream computation on $D$. The corresponding Concrete relaxation is to take $X \sim \Concrete(\alpha, \lambda)$ for some fixed temperature $\lambda \in (0, \infty)$ and set $\tilde{Y}  = C X$. For the binary case, this amounts to simply sampling $U \sim \Uniform(0,1)$ and taking $Y = 2H(\log U - \log(1-U) + \log \alpha) - 1$. The corresponding Binary Concrete relaxation is $\tilde{Y} = 2\sigma((\log U - \log(1-U) + \log \alpha)/\lambda) - 1$.

\subsection{Bias Initialization}
All biases were initialized to 0 with the exception of the biases in the prior decoder distribution over the 784 or 392 observed units. These were initialized to the logit of the base rate averaged over the respective dataset (MNIST or Omniglot).

\subsection{Centering}
We also found it beneficial to center the layers of the inference network during training. The activity in $(-1, 1)^d$ of each stochastic layer  was centered during training by maintaining a exponentially decaying average with rate 0.9  over minibatches. This running average was subtracted from the activity of the layer \emph{before} it was updated. Gradients did not flow throw this computation, so it simply amounted to a dynamic offset. The averages were \emph{not} updated during the evaluation. 

\subsection{Hyperparameter Selection}
All models were initialized with the heuristic of \citet{glorot2010understanding} and optimized using Adam \citep{kingma2014adam} with parameters $\beta_1 = 0.9, \beta_2 = 0.999$ for $10^{7}$ steps on minibatches of size 64. Hyperparameters were selected on the MNIST dataset by grid search taking the values that performed best on the validation set. Learning rates were chosen from $\{10^{-4}, 3\cdot10^{-4}, 10^{-3}\}$ and weight decay from $\{0, 10^{-2}, 10^{-1}, 1\}$. Two sets of hyperparameters were selected, one for linear models and one for non-linear models. The linear models' hyperparameters were selected with the 200H--200H--784V density model on the $\mathcal{L}_5(\theta, \phi)$ objective. The non-linear models' hyperparameters were selected with the 200H$\sim$200H$\sim$784V density model on the $\mathcal{L}_5(\theta, \phi)$ objective. For density estimation, the Concrete relaxation hyperparameters were (weight decay = 0, learning rate = $3\cdot10^{-4}$) for linear and (weight decay = 0, learning rate = $10^{-4}$) for non-linear. For structured prediction Concrete relaxations used  (weight decay = $10^{-3}$, learning rate = $3\cdot10^{-4}$). 

In addition to tuning learning rate and weight decay, we tuned temperatures for the Concrete relaxations on the density estimation task. We found it valuable to have different values for the prior and posterior distributions, see Eq. \ref{eq:tempchoose}. In particular, for binary we found that (prior $\lambda_2 = 1/2$, posterior $\lambda_1=2/3$) was best, for 4-ary we found (prior $\lambda_2 = 2/3$, posterior $\lambda_1=1$) was best, and (prior $\lambda_2 = 2/5$, posterior $\lambda_1=2/3$) for 8-ary.
No temperature annealing was used. For structured prediction we used just the corresponding posterior $\lambda_1$ as the temperature for the whole graph, as there was no variational posterior.

We performed early stopping when training with the score function estimators (VIMCO/NVIL) as they were much more prone to overfitting.

\clearpage

\section{Extra Results}
\label{appendix:extraresults}

\begin{table}[h]
\begin{center}
{\footnotesize
\begin{tabular}{@{}l l c c c c@{}}
       &   &  \multicolumn{2}{c}{MNIST NLL} & \multicolumn{2}{c}{Omniglot NLL} \\
\cmidrule(r){3-4} \cmidrule(l){5-6} 
    & $m$ & Test & Train & Test & Train \\
\toprule
\multirow{2}{*}{\shortstack[l]{binary\\(240H\\$\sim$784V)}} & 1   & 91.9 & 90.7 & 108.0 & 102.2 \\
                                                                         & 5  & 89.0 & 87.1 & 107.7 & 100.0 \\
                                                                         & 50  & 88.4 &  85.7 & 109.0 & 99.1 \\
\midrule
\multirow{2}{*}{\shortstack[l]{4-ary\\(240H\\$\sim$784V)}} & 1   & 91.4 & 89.7 & 110.7 & 1002.7 \\
                                                                         & 5  & 89.4 & 87.0 & 110.5 & 100.2 \\
                                                                         & 50  & 89.7 & 86.5 & 113.0 & 100.0 \\
\midrule
\multirow{2}{*}{\shortstack[l]{8-ary\\(240H\\$\sim$784V)}} & 1   & 92.5 &89.9 & 119.61 & 105.3\\
                                                                         & 5  & 90.5 & 87.0 & 120.7 & 102.7 \\
                                                                         & 50  & 90.5 & 86.7 & 121.7& 101.0\\
\midrule
\multirow{2}{*}{\shortstack[l]{binary\\(240H$\sim$240H\\$\sim$784V)}} & 1   &87.9&86.0&106.6 & 99.0\\
                                                                         & 5  &86.6&83.7&106.9&97.1\\
                                                                         & 50  &86.0&82.7&108.7&95.9\\
\midrule
\multirow{2}{*}{\shortstack[l]{4-ary\\(240H$\sim$240H\\$\sim$784V)}} & 1   & 87.4 & 85.0&106.6 & 97.8\\
                                                                         & 5  & 	86.7 & 83.3 &108.3&97.3\\
                                                                         & 50  & 86.7 & 83.0 &109.4 &96.8\\
\midrule
\multirow{2}{*}{\shortstack[l]{8-ary\\(240H$\sim$240H\\$\sim$784V)}} & 1   & 88.2 &85.9&111.3&102.5\\
                                                                         & 5  & 87.4 & 84.6&110.5&100.5\\
                                                                         & 50  & 87.2 & 84.0&111.1&99.5\\
\bottomrule
\end{tabular}}
\end{center}
\caption{ Density estimation using Concrete relaxations with distinct arity of layers.}
\label{tbl:narydensity}
\end{table}

\clearpage

\section{Cheat Sheet}
\label{appendix:cheatsheet}

\begin{align*}
\sigma(x) &= \frac{1}{1 + \exp(-x)} \;\;\;\;\;\;\;\; \logsumexp_{k=1}^n \{x_k\} = \log \left(\sum_{k=1}^n \exp(x_k)\right)\\
\log\Delta^{n-1} &= \left\{x \in \R^n \given x_k \in (-\infty,0),\  \logsumexp_{k=1}^n \{x_k\} = 0\right\}\\
\end{align*}

\begin{table}[h]
\begin{adjustbox}{center}
\begin{tabular}{@{}ll ll ll ll@{}}
Distribution and Domains & Reparameterization/How To Sample & Mass/Density \\ 
\toprule
 \multirow{3}{*}{\shortstack[l]{$G \sim \Gumbel$ \\ $G \in \R$}}  &\multirow{3}{*}{$G \overset{d}{=} -\log(-\log(U))$} & \multirow{3}{*}{$\exp(-g - \exp(-g))$} \\ 
  & &  \\ 
 & &  \\ 
\midrule
 \multirow{3}{*}{\shortstack[l]{$L \sim \Logistic$ \\ $L \in \R$}}  & \multirow{3}{*}{$\displaystyle L \overset{d}{=} \log(U) - \log(1-U)$} & \multirow{3}{*}{$\displaystyle\frac{\exp(- l)}{(1+ \exp(- l))^2}$} \\ 
 & &  \\ 
 & &  \\ 
\midrule
\multirow{4}{*}{\shortstack[l]{$X\sim \Logistic(\mu, \lambda)$ \\ $\mu \in \R$ \\ $\lambda \in (0, \infty)$}} & \multirow{4}{*}{$\displaystyle X \overset{d}{=} \frac{L + \mu}{\lambda}$}  & \multirow{4}{*}{$\displaystyle\frac{\lambda \exp(-\lambda x + \mu)}{(1+ \exp(-\lambda x + \mu))^2}$} \\ 
 & &  \\ 
 & &  \\ 
 & &  \\ 
\midrule
 \multirow{4}{*}{\shortstack[l]{$X \sim \Bernoulli(\alpha)$ \\ $X \in \{0,1\}$ \\ $\alpha \in (0, \infty)$}}  & \multirow{4}{*}{$\displaystyle X \overset{d}{=} \begin{cases}
      1 & \text{if $L + \log \alpha \geq 0$} \\
      0 & \text{otherwise}
    \end{cases}$} & \multirow{4}{*}{$\displaystyle
      \frac{\alpha}{1 + \alpha} $ if $x=1$} \\ 
 & &  \\ 
 & &  \\ 
 & &  \\ 
\midrule
 \multirow{5}{*}{\shortstack[l]{$X \sim \BinaryConcrete(\alpha, \lambda)$ \\ $X \in (0,1)$ \\ $\alpha \in (0, \infty)$ \\ $\lambda \in (0, \infty)$}}  & \multirow{5}{*}{$\displaystyle X \overset{d}{=} \sigma((L + \log \alpha)/\lambda)$} & \multirow{5}{*}{$\displaystyle \frac{\lambda \alpha x^{-\lambda - 1}(1-x)^{-\lambda - 1}}{(\alpha x^{-\lambda} + (1-x)^{-\lambda})^2}$} \\ 
 & &  \\ 
 & &  \\ 
 & &  \\ 
 & &  \\ 
\midrule
\multirow{5}{*}{\shortstack[l]{$X \sim \Discrete(\alpha)$\\ $X \in \{0,1\}^n$\\ $\sum_{k=1}^n X_k = 1$\\ $\alpha \in (0, \infty)^n$}} & \multirow{5}{*}{$\displaystyle X_k \overset{d}{=} \begin{cases}
      1 & \text{if $\log \alpha_k + G_k > \log \alpha_i + G_i$ for $i \neq k$} \\
      0 & \text{otherwise}
    \end{cases}$} & \multirow{5}{*}{\shortstack[l]{$\displaystyle\frac{\alpha_k}{\sum_{i=1}^n \alpha_i}$ if $x_k = 1$}} \\ 
 & &  \\ 
 & &  \\ 
 & &  \\ 
 & &  \\ 
\midrule
\multirow{5}{*}{\shortstack[l]{$X \sim \Concrete(\alpha, \lambda)$ \\ $X \in \Delta^{n-1}$ \\ $\alpha \in (0, \infty)^n$ \\ $\lambda  \in (0, \infty)$}} & \multirow{5}{*}{$\displaystyle X_k \overset{d}{=} \frac{\exp((\log \alpha_k + G_k)/\lambda)}{\sum_{i=1}^n \exp((\log \alpha_k + G_i)/\lambda)}$} & \multirow{5}{*}{\shortstack[l]{$\displaystyle \frac{(n-1)!}{\lambda^{-(n-1)}} \prod_{k=1}^n \frac{\alpha_k x_k^{-\lambda - 1}}{\sum_{i=1}^n \alpha_i x_i^{-\lambda}}$}} \\ 
 & &  \\ 
 & &  \\ 
 & &  \\ 
 & &  \\ 
\midrule
\multirow{5}{*}{\shortstack[l]{$X \sim \ExpConcrete(\alpha, \lambda)$ \\ $X \in \log\Delta^{n-1}$ \\ $\alpha \in (0, \infty)^n$ \\ $\lambda  \in (0, \infty)$}} & \multirow{5}{*}{$\displaystyle X_k \overset{d}{=} \frac{\log \alpha_k + G_k}{\lambda} - \logsumexp_{i=1}^n \left\{ \frac{\log \alpha_i + G_i}{\lambda}\right\}$} & \multirow{5}{*}{\shortstack[l]{$\displaystyle \frac{(n-1)!}{\lambda^{-(n-1)}} \prod_{k=1}^n \frac{ \alpha_k\exp(-\lambda x_k)}{\sum_{i=1}^n  \alpha_i\exp(-\lambda x_i)}$}} \\ 
 & &  \\ 
 & &  \\ 
 & &  \\ 
 & &  \\ 
\bottomrule
\end{tabular}
\end{adjustbox}
\caption{Cheat sheet for the random variables we use in this work. Note that some of these are atypical parameterizations, particularly the Bernoulli and Logistic random variables. The table only assumes that you can sample uniform random numbers $U \sim \Uniform(0,1)$. From there on it may define random variables and reuse them later on. For example, $L \sim \Logistic$ is defined in the second row, and after that point $L$ represents a Logistic random variable that can be replaced by $\log U - \log(1-U)$. Whenever random variables are indexed, e.g. $G_k$, they represent separate independent calls to a random number generator.}
\label{tbl:cheatsheet}
\end{table}